\documentclass[lettersize,journal]{IEEEtran}
\usepackage{amsmath,amsfonts}
\usepackage{algorithmic}
\usepackage{algorithm}
\usepackage{array}
\usepackage[caption=false,font=normalsize,labelfont=sf,textfont=sf]{subfig}
\usepackage{textcomp}
\usepackage{stfloats}
\usepackage{url}
\usepackage{verbatim}
\usepackage{graphicx}
\usepackage{cite}
\usepackage{wrapfig}
\usepackage{makecell}
\usepackage{multirow}
\usepackage{xcolor}  
\usepackage{hhline}
\usepackage{colortbl}  
\newcommand{\etal}{\textit{et al.}}
\newcommand{\eg}{\textit{e.g.}}
\usepackage{amssymb}
\usepackage{pifont}

\begin{document}
\title{Boosting Semi-Supervised 2D Human Pose Estimation by Revisiting Data Augmentation and Consistency Training}

\author{Huayi Zhou, Mukun Luo, Fei Jiang, Yue Ding, Hongtao Lu~\IEEEmembership{Member,~IEEE} and Kui Jia~\IEEEmembership{Member,~IEEE}
\thanks{H. Zhou and K. Jia (the corresponding author) are with School of Data Science, The Chinese University of Hong Kong, Shenzhen (e-mail: zhouhuayi@cuhk.edu.cn; kuijia@cuhk.edu.cn). F. Jiang is with the Chongqing Academy of Science and Technology (e-mail: fjiang@mail.ecnu.edu.cn). M. Luo, Y. Ding, and H. Lu are with Department of Computer Science and Engineering, Shanghai Jiao Tong University, Shanghai 200240, China (e-mail: luomukun@sjtu.edu.cn; dingyue@sjtu.edu.cn; htlu@sjtu.edu.cn).}
\thanks{Manuscript received February 19, 2025; revised May 16, 2025.}}

\markboth{Journal of \LaTeX\ Class Files,~Vol.~14, No.~8, August~2025}%
{Shell \MakeLowercase{\textit{et al.}}: A Sample Article Using IEEEtran.cls for IEEE Journals}


\maketitle
\begin{abstract}
The 2D human pose estimation (HPE) is a basic visual problem. However, its supervised learning requires massive keypoint labels, which is labor-intensive to collect. Thus, we aim at boosting a pose estimator by excavating extra unlabeled data with semi-supervised learning (SSL). Most previous SSHPE methods are consistency-based and strive to maintain consistent outputs for differently augmented inputs. Under this genre, we find that SSHPE can be boosted from two cores: advanced data augmentations and concise consistency training ways. Specifically, for the first core, we discover the synergistic effects of existing augmentations, and reveal novel paradigms for conveniently producing new superior HPE-oriented augmentations which can more effectively add noise on unlabeled samples. We can therefore establish paired easy-hard augmentations with larger difficulty gaps. For the second core, we propose to repeatedly augment unlabeled images with diverse hard augmentations, and generate multi-path predictions sequentially for optimizing multi-losses in a single network. This simple and compact design is interpretable, and easily benefits from newly found augmentations. Comparing to state-of-the-art SSL approaches, our method brings substantial improvements on public datasets. And we extensively validate the superiority and versatility of our approach on conventional human body images, overhead fisheye images, and human hand images. The code is released in \url{https://github.com/hnuzhy/MultiAugs}.
\end{abstract}


\section{Introduction}\label{intro}

\begin{figure*}[t]
	\centering
	\subfloat[{\footnotesize Single-Network}]{\includegraphics[width=0.200\textwidth]{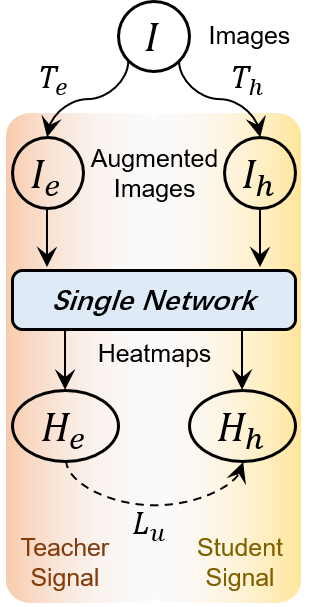}
	\label{OldA}}
	\subfloat[{\footnotesize Dual-Network}]{\includegraphics[width=0.315\textwidth]{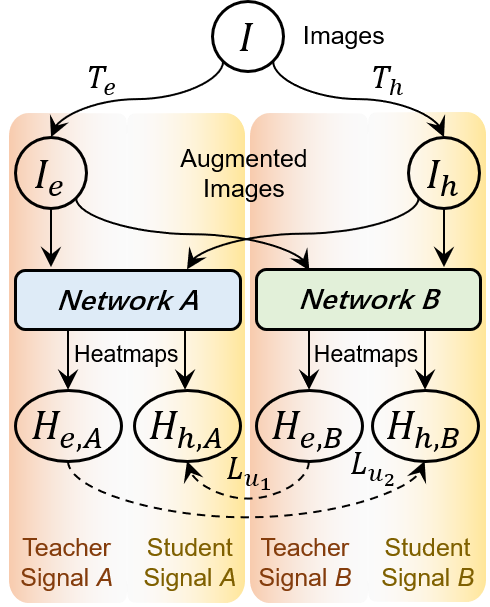}
	\label{OldB}}
	\subfloat[{\footnotesize Triple-Network}]{\includegraphics[width=0.475\textwidth]{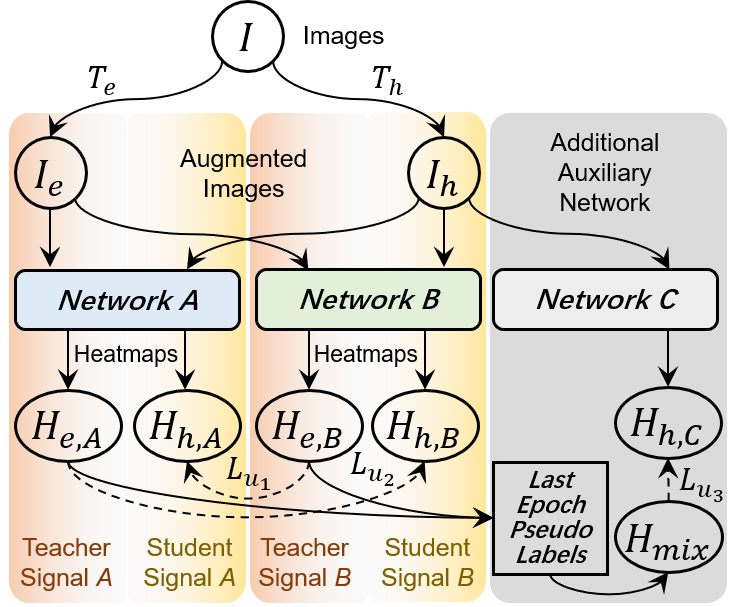}
	\label{OldC}}
	\caption{Frameworks of existing semi-supervised human pose estimation (SSHPE) methods including {(a)} Single-Network and {(b)} Dual-Network which is originally proposed by \cite{xie2021empirical}, and {(c)} Triple-Network that is proposed by \cite{huang2023semi}.}
	\vspace{-10pt}
	\label{frameworkOld}
\end{figure*}

\IEEEPARstart{T}{he} 2D human pose estimation (HPE) aims to detect and represent human parts as sparse 2D keypoint locations in RGB images. It is the basis of many visual tasks such as action recognition \cite{yan2018spatial, duan2022revisiting}, person re-identification \cite{zhao2017spindle, sarfraz2018pose}, 3D pose lifting \cite{nie2023lifting, dabhi20243d} and 3D human shape regression \cite{pavlakos2018learning, pavlakos2019expressive}. Modern data-driven HPE has been substantially improved by generous deep supervised learning approaches \cite{cao2017realtime, cheng2020higherhrnet, xu2022vitpose, yang2023explicit, tan2024diffusionregpose}. This greatly benefits from the collection and annotation of many large-scale public HPE datasets \cite{andriluka20142d, lin2014microsoft, wu2019large}. However, compared to image classification and detection tasks requiring plain labels, obtaining accurate 2D keypoints from massive images is laborious and time-consuming. To this end, some researches \cite{xie2021empirical, moskvyak2021semi, wang2022pseudo, huang2023semi, yu2024denoising} try to alleviate this problem by introducing the semi-supervised 2D human pose estimation (SSHPE), which can subtly leverage extensive easier obtainable yet unlabeled 2D human images in addition to partial labeled data to improve performance. Although existing methods \cite{xie2021empirical, huang2023semi} have improved the accuracy of SSHPE task, they overlooked two fundamental questions:

\textsf{Q1}: \textbf{How to judge the discrepancy of unsupervised data augmentations with different difficulty levels?}
As shown in Fig.~\ref{OldA}, for a batch of unlabeled images $\mathbf{I}$, its easy augmentation $\mathbf{I}_e$ and hard augmentation $\mathbf{I}_h$ are generated separately in \cite{xie2021empirical}. Then, predicted heatmap $\mathbf{H}_e$ of $\mathbf{I}_e$ is used as a pseudo label to teach the network to learn the harder counterpart $\mathbf{I}_h$ with its yielded heatmap $\mathbf{H}_h$. \cite{xie2021empirical} finds that the large gap between two augmentations $(\mathbf{I}_e, \mathbf{I}_h)$ matters. Essentially, this is a pursuit for more advanced data augmentations in SSHPE. To rank augmentations of different difficulty levels, \cite{xie2021empirical} observes precision degradation of a pretrained model by testing it on a dataset after corresponding augmentation. However, we declare that this manner is not rigorous. An obvious counter-case is that over-augmented samples approaching noise will get worst evaluation results, but such hard augmentations are meaningless. Contrastingly, we deem that a persuasive ranking requires independent training for each augmentation. We answer this question in detail in Sec.~\ref{studyAugs}.

\textsf{Q2}: \textbf{How to generate multiple unsupervised signals for consistency training efficiently and concisely?}
Previous work \cite{xie2021empirical} proposes to use a Single-Network to perform the unsupervised consistency training on the easy-hard pair $(\mathbf{I}_e, \mathbf{I}_h)$. It also gives a more complicated Dual-Network as in Fig.~\ref{OldB} for cross-training of two easy-hard pairs. SSPCM \cite{huang2023semi} even constructs a Triple-Network as in Fig.~\ref{OldC} for interactive training of three easy-hard pairs. This pattern of adding more networks with the increase of unsupervised signal pairs can certainly bring gains. But this is cumbersome and will decelerate the training speed proportionally. In practice, considering that augmentations are always performed on the same input, we can repeatedly augment $\mathbf{I}$ multiple times with $n$ diverse hard augmentations, and generate multi-path easy-hard pairs $\{(\mathbf{I}_e, \mathbf{I}_{h_1}), (\mathbf{I}_e, \mathbf{I}_{h_2}), ..., (\mathbf{I}_e, \mathbf{I}_{h_n})\}$. In this way, we can use only one single network (refer Fig.~\ref{NewA}) to optimize $n$ pairs of losses. This is also applicable to dual networks (refer Fig.~\ref{NewB}). We discuss this question in detail in Sec.~\ref{studyMulti}. 

In this paper, we mainly revisit \textsf{Q1} and \textsf{Q2} to boost SSHPE. For \textsf{Q1}, after properly ranking existing basic augmentations \cite{devries2017improved, zhang2018mixup, yun2019cutmix, cubuk2018autoaugment, cubuk2020randaugment}, we naturally try to extend it to discover new strong augmentations through reasonable sequential combinations inspired by the AutoAug families \cite{cubuk2018autoaugment, lim2019fast, hataya2020faster, zheng2022deep}. Rather than trivial enumeration, we notice the \textbf{synergistic effects} between different augmentations, and reveal novel paradigms for easily generating superior combinations. Our paradigms for the SSHPE task contain three principles: \textsf{(P1)} \textit{Do not combine MixUp-related augmentations.} \textsf{(P2)} \textit{Try to utilize the synergistic effects.} \textsf{(P3)} \textit{Do not over-combine too many augmentations.} These principles have favorable interpretability, and bypass painstaking designs in other advanced augmentations \cite{hendrycks2020augmix, muller2021trivialaugment, zheng2022deep, pinto2022using, liu2022automix, han2022you}. For \textsf{Q2}, we quantitatively validated the superiority of multi-path design over commonly used heatmaps fusion \cite{radosavovic2018data} and confidence masking \cite{xie2020unsupervised, huang2023semi}. Combining it with newly found advanced augmentations, our Single-Network based approach can surpass the original Dual-Network \cite{xie2021empirical} and come close to SSPCM using a Triple-Network.

In summary, our contributions are three-folds: 
\begin{itemize}
\item We comprehensively evaluated difficulty levels of existing data augmentations suitable for the SSHPE task, validated their synergistic effects by properly combining different basic augmentations, and presented novel combination paradigms which are intuitively interpretable.
\item We proposed to easily generate multi-path predictions of strongly augmented samples in separate for training only one single model, rather than adding more auxiliary networks. Therefore, we can optimize multiple unsupervised losses efficiently and concisely, and benefit from distinct superior augmentations.
\item We achieved new state-of-the-art results on public SSHPE benchmarks (ranging from conventional human body images to overhead fisheye images or conventional human hand images) with less training time and parameters under same settings of previous methods.
\end{itemize}

Our paper is organized as follows. In Sec.~\ref{relwork}, we introduce various related works, including semi-supervised learning (SSL), semi-supervised human pose estimation (SSHPE), and unsupervised data augmentation techniques. In Sec.~\ref{empstu}, we first give a formal definition of SSHPE. Then, we focus on answering the raised two core questions in the form of empirical studies, along with theoretical analysis of the effectiveness of advanced augmentations. In Sec.~\ref{methodMAs}, we explain the overall training framework of the proposed MultiAugs using the Single-Network or Dual-Network, which has wide adaptability and versatility. Finally, in Sec.~\ref{exps}, we present a large number of experimental results, including quantitative and qualitative comparisons of our MultiAugs with other methods on multiple HPE datasets, as well as sufficient ablation studies.

\section{Related Work}\label{relwork}


\subsection{Semi-Supervised Learning (SSL)}
SSL is originated in the classification task by exploiting a small set of labeled data and a large set of unlabeled data. It can be categorized into pseudo-label (PL) based \cite{radosavovic2018data, oliver2018realistic, xie2020self, sohn2020fixmatch, guo2022class, wang2023freematch} and consistency-based \cite{laine2016temporal, tarvainen2017mean, berthelot2019mixmatch, xie2020unsupervised, zhang2021flexmatch, gui2023enhancing, huang2024interlude}. PL-based methods iteratively add unlabeled images into the training data by pseudo-labeling them with a pretrained or gradually enhanced model. It needs to find suitable thresholds to mask out uncertain samples with low-confidence, which is a crucial yet tricky issue. Consistency-based methods enforce model outputs to be consistent when its input is randomly perturbed. They have shown to work well on many tasks. For example, MixMatch \cite{berthelot2019mixmatch} combines the consistency regularization with the entropy minimization to obtain confident predictions. FixMatch \cite{sohn2020fixmatch} utilizes a weak-to-strong consistency regularization and integrates the pseudo-labeling to leverage unlabeled data. FlexMatch \cite{zhang2021flexmatch} and FreeMatch \cite{wang2023freematch} adopt the curriculum learning and adaptive thresholding based on FixMatch, respectively. CRMatch \cite{fan2023revisiting} and SAA \cite{gui2023enhancing} try to design strategies and augmentations to enhance the consistency training. These SSL methods give us primitive inspirations.

\subsection{Semi-Supervised Human Pose Estimation (SSHPE)}
This is relatively less-studied comparing to other visual tasks like classification and object detection. A few SSHPE methods are based on pseudo labeling \cite{wang2022pseudo, springstein2022semi, yu2024denoising} or consistency training \cite{xie2021empirical, moskvyak2021semi, li2023scarcenet, huang2023semi, han2022learning}. SSKL \cite{moskvyak2021semi} designs a semantic keypoint consistency constraint to learn invariant representations of same keypoints. It is evaluated on small-scale HPE datasets MPII \cite{andriluka20142d} and LSP \cite{johnson2011learning}, instead of the larger COCO \cite{lin2014microsoft}. Following it, PLACL \cite{wang2022pseudo} introduces the curriculum learning by auto-selecting dynamic thresholds for producing pseudo-labels via reinforcement learning. Inspired by co-training \cite{qiao2018deep} and dual-student \cite{ke2019dual}, Dual-Network \cite{xie2021empirical} points out the typical collapsing problem, and proposes the easy-hard augmentation pair on the same input to imitate teacher-student signals without relying on Mean Teacher \cite{tarvainen2017mean}. SSPCM \cite{huang2023semi} extends the Dual into Triple by adding an auxiliary teacher for interactive training in multi-steps. It designs a handcrafted pseudo-label correction based on the predicted position inconsistency of two teachers, and has achieved SOTA performances. Still based on Dual-Network, Pesudo-HMs \cite{yu2024denoising} utilizes the cross-student uncertainty to propose a threshold-refine procedure, which can select reliable pseudo-heatmaps as targets for learning from unlabeled data. While, in this paper, we revisit the less efficient consistency training in \cite{xie2021empirical, huang2023semi}, and propose to upgrade the Single-Network by exploring multi-path predictions. 

\subsection{Unsupervised Data Augmentations}
The UDA \cite{xie2020unsupervised} has emphasized and verified the key role of high-quality noise injection (\eg, data augmentations) in improving unsupervised consistency training. It utilizes advanced augmentations \cite{cubuk2018autoaugment, cubuk2020randaugment} to promote the SSL classification. Then, \cite{xie2021empirical} transfers the positive correlation between strong augmentations and SSL performance to the HPE field. It introduces a more advanced augmentation called Joint Cutout inspired by Cutout \cite{devries2017improved}. Similarly, SSPCM \cite{huang2023semi} provides a harder keypoints-sensitive augmentation Cut-Occlude inherited from CutMix \cite{yun2019cutmix}. In this paper, we thoroughly revisit existing data augmentations suitable to SSHPE, give a rank of their difficulty levels by controlled training, and produce simple paradigms for getting novel superior joint-related augmentations. We also compare them with other well-designed counterparts \cite{cubuk2020randaugment, muller2021trivialaugment, han2022you} to reveal our advantages.

\section{Empirical Studies}\label{empstu}

{\bf Problem Definition}:
The task of 2D HPE is to detect $k$ body joints in an image $\mathbf{I}\in\mathbb{R}^{h\times w\times 3}$. The state-of-the-art methods \cite{xiao2018simple, sun2019deep} tend to estimate $k$ Gaussian heatmaps $\mathbf{H}\in\mathbb{R}^{\frac{h}{s}\times \frac{w}{s}\times k}$ downsampled $s$ times. For inference, each keypoint is located by finding the pixel with largest value in its predicted heatmap. We denote the labeled and unlabeled training sets as $\mathcal{D}^l\!=\!\{(\mathbf{I}^l_i, \mathbf{H}^l_i)\}|^N_{i=1}$ and $\mathcal{D}^u\!=\!\{\mathbf{I}^u_i\}|^M_{i=1}$, respectively. Here, the $\mathbf{I}^l_i$ or $\mathbf{I}^u_i$ means a labeled or unlabeled image sample, respectively. And $N$ or $M$ is the total number of image samples. The $\mathbf{H}^l$ are ground-truth heatmaps generated using 2D keypoints. For supervised training of the network $f$, we calculate the MSE loss:
\begin{equation}  
	\mathcal{L}_s = \mathbb{E}_{\mathbf{I}\in\mathcal{D}^l} || f(T_e(\mathbf{I})) - T_e(\mathbf{H}) ||^2, ~
	\label{losssup}
\end{equation}
where $T_e$ represents an easy affine augmentation including a random rotation angle from $[-30^\circ, 30^\circ]$ and scale factor from $[0.75, 1.25]$ (denoted as $T_{A30}$). For unlabeled images, we calculate the unsupervised consistency loss:
\begin{equation}  
	\mathcal{L}_u = \mathbb{E}_{\mathbf{I}\in\mathcal{D}^u} || T_{e \rightarrow h}(f(T_e(\mathbf{I}))) - f(T_h(\mathbf{I})) ||^2, ~
	\label{lossunsup}
\end{equation}
where $T_h$ is a harder augmentation with strong perturbations than affine-based $T_e$. The $T_{e \rightarrow h}$ means a known affine transformation on heatmaps if $T_h$ contains additional rotation and scaling operations. In this way, we can obtain a paired \textit{easy-hard} augmentations $(\mathbf{I}_e, \mathbf{I}_h)\!=\!(T_e(\mathbf{I}), T_h(\mathbf{I}))$ for generating corresponding teacher signals and student signals. During training, we should stop gradients propagation of teacher signals to avoid collapsing. Next, we answer two questions \textsf{Q1} and \textsf{Q2} by extensive empirical studies in Sec.~\ref{studyAugs} and Sec.~\ref{studyMulti}, respectively. After that, we provide a theoretical perspective of pursuiting stronger augmentations in Sec.~\ref{studyAnaly}.

\subsection{Paradigms of Generating Superior Augmentations}\label{studyAugs} 

\begin{figure}
	\centering
	\includegraphics[width=\columnwidth]{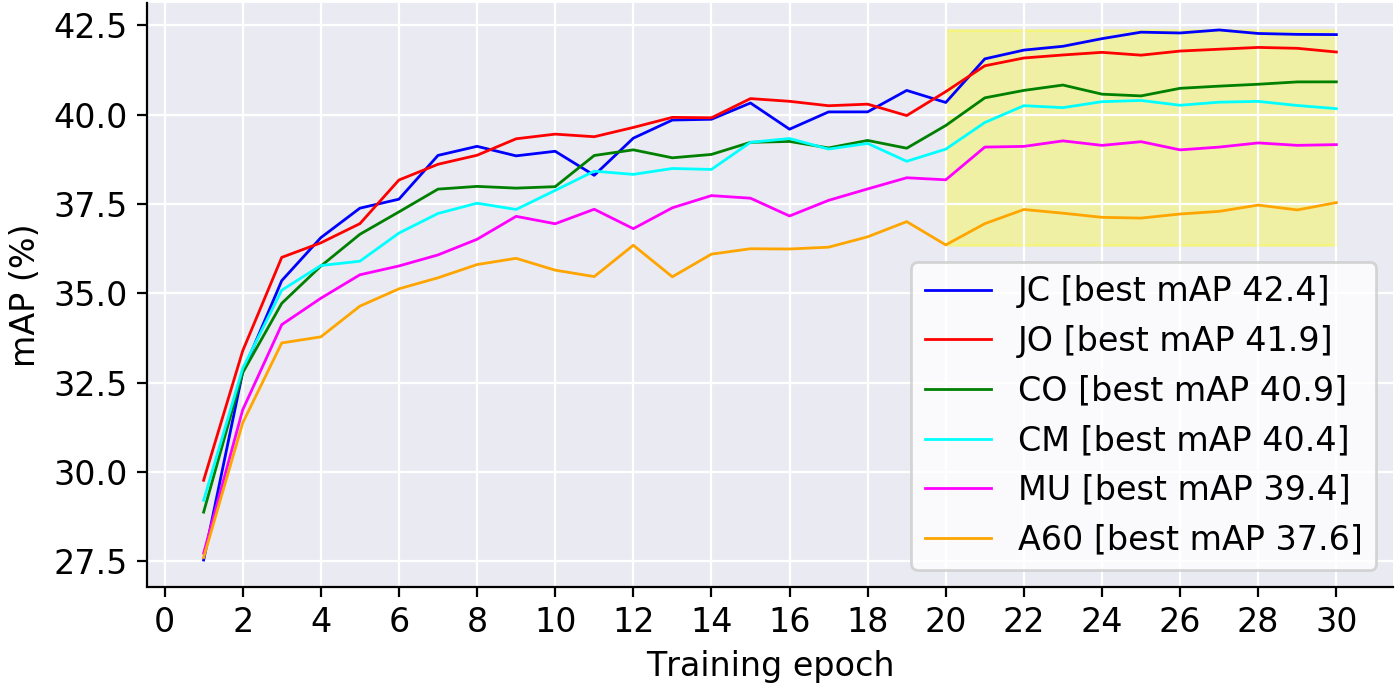}
	\vspace{-15pt}
	\caption{Comparison of applying different easy-hard pairs for training a Single-Network model as in Fig.~\ref{OldA}. We can rank these six augmentations indisputably based on either best mAP results or distinct convergence curves.}
	\label{AugsRanking}
	\vspace{-10pt}
\end{figure}

\subsubsection{Ranking of Basic Augmentations}
The core of the easy-hard pair paradigm $(\mathbf{I}_e, \mathbf{I}_h)$ is an advanced augmentation. To this end, Dual-Network \cite{xie2021empirical} and SSPCM \cite{huang2023semi} propose pseudo keypoint-based augmentations Joint Cutout ($T_{JC}$) and Joint Cut-Occlude ($T_{JO}$), respectively. They also reach a similar yet crude conclusion about difficulty levels of existing augmentations: $\{T_{JO}, T_{JC}\}\!>\!\{T_{RA}, T_{CM}, T_{CO}, T_{MU}, T_{A60}\}$, where $T_{RA}$, $T_{CM}$, $T_{CO}$ and $T_{MU}$ are RandAugment \cite{cubuk2020randaugment}, CutMix \cite{yun2019cutmix}, Cutout \cite{devries2017improved} and Mixup \cite{zhang2018mixup}, respectively. The $T_{A60}$ consists of two $T_{A30}$. We give them a new ranking by conducting more rigorous trainings one-by-one. The $T_{RA}$ is removed for it contains repetitions with $T_{CO}$ and $T_{A60}$. As shown in Fig.~\ref{AugsRanking}, we divide the rest by their distinguishable gaps into four levels: $\{T_{JC}, T_{JO}\}\!>\!\{T_{CO}, T_{CM}\}\!>\!\{T_{MU}\}\!>\!\{T_{A60}\}$.

\begin{figure}
	\centering
	\includegraphics[width=\columnwidth]{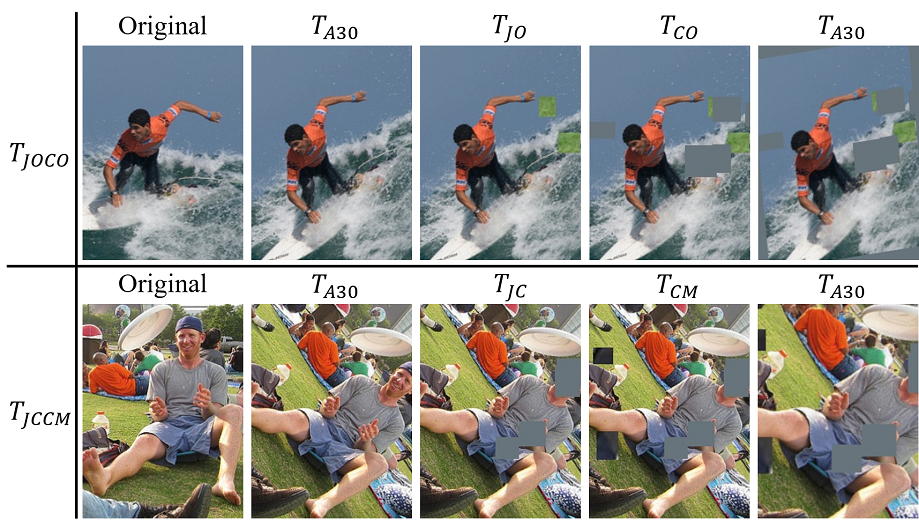}
	\vspace{-15pt}
	\caption{Illustrations of superior combinations $T_{JOCO}$ and $T_{JCCM}$. Either of them is a sequential operations of ready-made collaborative augmentations. $T_{JO}$ and $T_{CM}$ need extra patches cropped from other images which are not displayed.}
	\label{AugsPlus}
	\vspace{-10pt}
\end{figure}

\subsubsection{Synergy between Augmentations}
Then, instead of laboriously designing stronger augmentations, we consider to conduct two or more augmentations in sequence to obtain superior combinations conveniently. This idea is feasible because it essentially belongs to the AutoAug families \cite{cubuk2018autoaugment, lim2019fast, hataya2020faster, zheng2022deep}. Instead of auto-searching, we expect to find some heuristic strategies for the HPE task. In fact, after performing joint-related $T_{JO}$ or $T_{JC}$ on one image, we can continue to perform some joint-unrelated augmentations such as $T_{CM}$, $T_{CO}$ and $T_{MU}$ on random areas. As shown in Fig.~\ref{AugsPlus}, applying $T_{JOCO}$ (a $T_{CO}$ after $T_{JO}$) or $T_{JCCM}$ (a $T_{CM}$ after $T_{JC}$) will bring harder samples for generating more effective student signals, but not destroy the semantic information visually. We call this discovery the \textbf{synergistic effect} between different augmentations. The $T_{A60}$ can server as an essential factor in any $T_h$ for keeping the geometric diversity.

\subsubsection{Selection of Augmentations Combination}\label{SelAugCom}
Now, if selecting from the rest five basic augmentations, there are up to 26 choices $(2^5-\binom{5}{0}-\binom{5}{1})$. The optimal combinations are still time-consuming to acquire. Fortunately, not arbitrary number or kind of augmentations are collaborative. We intuitively summarize three simplistic principles. \textsf{(P1)} A global $T_{MU}$ does not make sense for the HPE task. \textsf{(P2)} Stacking augmentations with the similar perturbation type (e.g., $T_{JO}\!\sim\!T_{CM}$ and $T_{JC}\!\sim\!T_{CO}$) or difficulty level (e.g., $T_{JO}\!\sim\!T_{JC}$ and $T_{CM}\!\sim\!T_{CO}$) may not bring significant gain. \textsf{(P3)} Adding too many augmentations (e.g., three or four) will be profitless or even harmful for seriously polluting the image. We thus nominate the most likely superior combinations: $T_{JOCO}$ and $T_{JCCM}$. A case of setting $T_e$ as $T_{A30}$ and $T_h$ as $T_{JOCO}$ for getting corresponding easy teacher signals $\mathbf{H}_e$ and hard student signals $\mathbf{H}_h$ is shown as below:
\begin{equation}  
\begin{aligned}
	&\mathbf{H}_e = T_{A30 \rightarrow A60}(f(\mathbf{I}_e)), \quad \mathbf{I}_e = T_{A30}(\mathbf{I}), \\~
	&\mathbf{H}_h = f(\mathbf{I}_h), \quad \mathbf{I}_h = T_{A30}(T_{CO}(T_{JO}(T_{A30}(\mathbf{I}) ) ) ),  ~
	\label{heatEH}
\end{aligned}
\end{equation}
where $T_{A60}$ is the default of $T_h$, and divided into two separate $T_{A30}$ for being compatible with $T_e$. To further verify the above intuitive principles, we follow the augmentations ranking way and conduct empirical studies on the performance of up to 13 selected representative combinations out of 26 choices.

As shown in Tab.~\ref{tabAugsPlusAS} and Fig.~\ref{AugsPlusAS}, we can examine three principles one-by-one: \textsf{(P1)} \textit{$T_{MU}$ often causes adverse effects for each combination.} Please refer paired combinations $c_1$-$c_3$, $c_2$-$c_4$, $c_1$-$c_8$ and $c_2$-$c_9$. Thus, we do not add it for finding superior combinations. \textsf{(P2)} \textit{Synergistic effects between augmentations do exist.} Please refer paired combinations $c_1$-$c_5$, $c_1$-$c_7$, $c_2$-$c_6$ and $c_2$-$c_7$. Especially, the $T_{JCJO}$ with two most advanced augmentations performs the worst among combinations \{$c_1$,$c_2$,$c_5$,$c_6$,$c_7$\}, which roundly reveals the harm of violating the principle of synergy. \textsf{(P3)} \textit{Do not overly combine too many augmentations.} Please refer paired combinations $c_1$-$c_{12}$, $c_2$-$c_{13}$, $c_1$-$c_{11}$ and $c_2$-$c_{10}$. Stacking more augmentations brings non-significant gains or even results in degradation. We attribute it to deviating from the rule of collaboration and possibly producing meaningless or difficult-to-recognize images. Based on these facts, we have sufficient reasons not to check the performance of rest combinations, and recommend two new strongest combinations $T_{JOCO}$ and $T_{JCCM}$.

\vspace{0.2cm}
\makeatletter\def\@captype{table}\makeatother
\hspace*{-0.3cm}\begin{minipage}{.23\textwidth}\footnotesize 
	\centering
	\caption{Best mAPs of different combinations.}
	\vspace{-5pt}
	\label{tabAugsPlusAS}
	\setlength{\tabcolsep}{4pt}
	\begin{tabular}{c|c|c}
	\Xhline{1.2pt}
	Id & Combination & mAP \\
	\Xhline{1.2pt}
	-- & $T_{MU}$ & 39.4 \\
	-- & $T_{CO}$ & 40.9 \\
	-- & $T_{CM}$ & 40.4 \\
	-- & $T_{JC}$ & 42.4 \\
	-- & $T_{JO}$ & 41.9 \\
	\hline\rowcolor{gray!20}
	$c_1$ & $T_{JC,CM}$ & 42.7 \\
	\rowcolor{gray!20}
	$c_2$ & $T_{JO,CO}$ & {\bf 43.7} \\
	$c_3$ & $T_{JC,MU}$ & 41.8 \\
	$c_4$ & $T_{JO,MU}$ & 42.1 \\
	$c_5$ & $T_{JC,CO}$ & 42.1 \\
	$c_6$ & $T_{JO,CM}$ & 42.5 \\
	$c_7$ & $T_{JC,JO}$ & 42.0 \\
	\hline
	$c_8$ & $T_{JC,CM,MU}$ & 42.0 \\
	$c_9$ & $T_{JO,CO,MU}$ & 42.8 \\
	$c_{10}$ & $T_{JC,JO,CO}$ & 41.7 \\
	$c_{11}$ & $T_{JC,JO,CM}$ & 42.3 \\
	$c_{12}$ & $T_{JC,CO,CM}$ & 42.8 \\
	$c_{13}$ & $T_{JO,CO,CM}$ & 42.8 \\
	\Xhline{1.2pt}
	\end{tabular}
\end{minipage}
\makeatletter\def\@captype{figure}\makeatother
\begin{minipage}{.25\textwidth} 
	\centering
	\includegraphics[width=\linewidth]{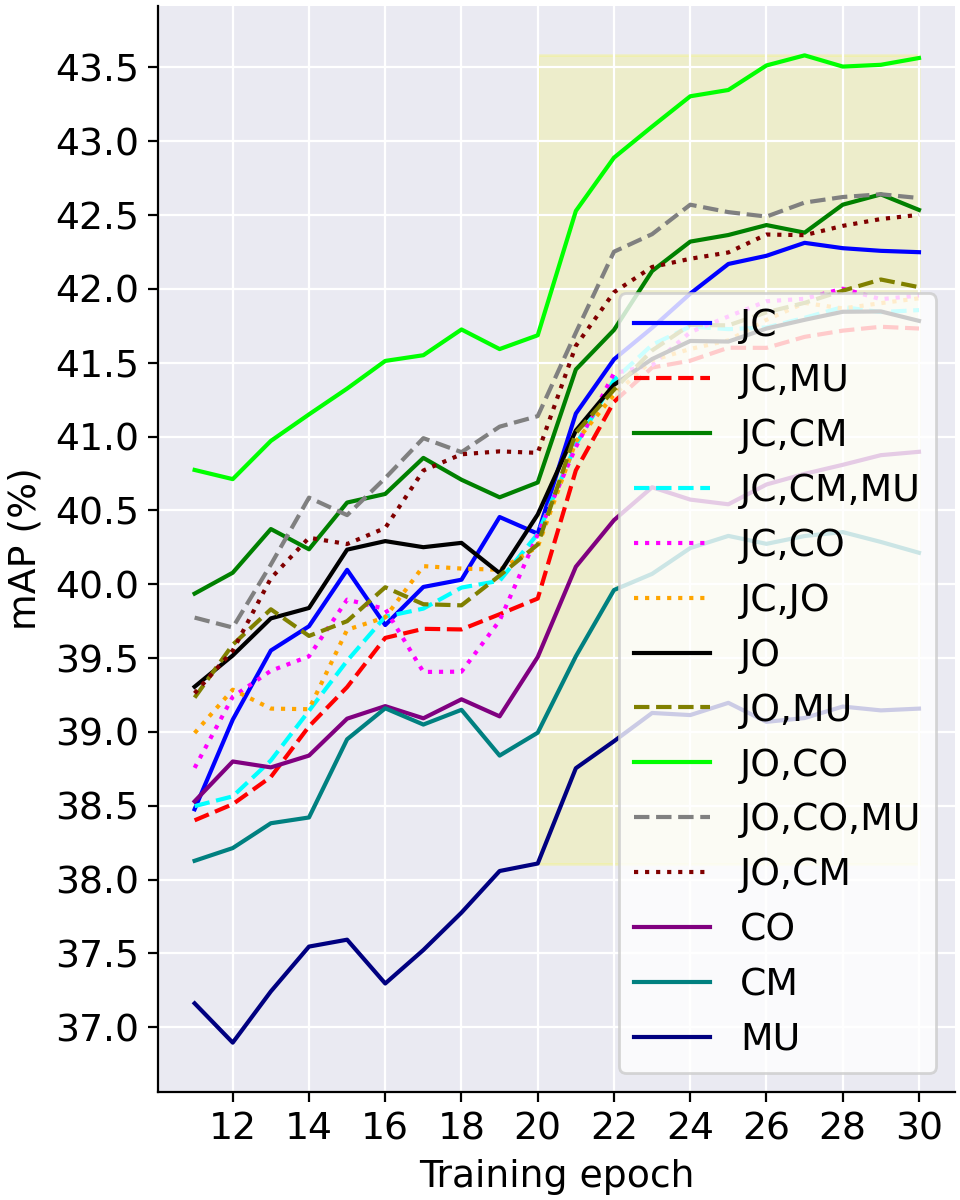} \\
	\vspace{-5pt}
	\caption{The corresponding convergence curves of combinations in Tab.~\ref{tabAugsPlusAS}.}
	\label{AugsPlusAS}
\end{minipage}
\vspace{0.1cm}

\subsection{Multi-path Consistency Losses}\label{studyMulti}

\subsubsection{Intuitive Motivation}
To further amplify the advantage of easy-hard augmentation, \cite{xie2021empirical} adopts two independent networks containing two easy-hard pairs for producing two consistency losses. SSPCM \cite{huang2023semi} continues this route by designing a Triple-Network with three easy-hard pairs. Meanwhile, some SSL methods construct multi-view inputs for unlabeled data without adding accompanying networks. For example, SwAV \cite{caron2020unsupervised} enforces the local-to-global consistency among a bag of views with different resolutions. ReMixMatch \cite{berthelot2020remixmatch} feeds multiple strongly augmented versions of an input into the model for training. Therefore, we wonder whether such a simple idea can also benefit the SSHPE task.

Specifically, rather than feeding a single hard augmentation $\mathbf{I}_h$ into the model, we independently yield $n$ strongly augmented inputs $\mathcal{I}^n\!=\!\{\mathbf{I}_{h_1}, \mathbf{I}_{h_2}, ..., \mathbf{I}_{h_n}\}$ from $\mathbf{I}$ by applying $n$ hard data augmentations $\mathcal{T}^n\!=\!\{T_{h_1}, T_{h_2}, ..., T_{h_n}\}$ accordingly. The augmentation set $\mathcal{T}^n$ is de-emphasized in order and non-deterministic, and will generate distinct multi-path augmented inputs $\mathcal{I}^n$. Then, we can calculate $n$-stream heatmaps $\mathcal{H}^n\!=\!\{f(T_{h_i}(\mathbf{I}_{h_i}))|^n_{i=1}\}$. This multi-path augmentation framework is illustrated in Fig.~\ref{frameworkNew}. For regularizing $n$ easy-hard pairs, we obtain multi-path consistency losses using Eq.~\ref{lossunsup} in separate, and optimize them jointly by applying multi-loss learning:
\begin{equation}  
	\mathcal{L}^*_u = \mathbb{E}_{\mathbf{H}_{h_i}\in\mathcal{H}^n}\sum\nolimits^n_{i=1} || \mathbf{H}_e - \mathbf{H}_{h_i} ||^2, ~
	\label{lossesunsup}
\end{equation}
where $\mathbf{H}_e$ and $\mathbf{H}_{h_i}$ are obtained as dissected in Eq.~\ref{heatEH}. The $\mathbf{H}_e$ keeps constant for each $\mathbf{H}_{h_i}$. Comparably, Data Distill \cite{radosavovic2018data} applies a single model to multiple transformations of unlabeled data to train a student model. Then, it ensembles predictions to obtain keypoint locations, and re-generates a pseudo heatmap for supervision. Differently, we argue that conducting a fusion on predicted heatmaps in SSHPE is harmful. We consider that there are always differences in the estimation of keypoint positions for each $\mathbf{I}_{h_i}$. It is an ill-posed problem to heuristically evaluate the consistency regularization contribution of each heatmap in pixel during ensemble. We will explain this in ablation studies Sec.~\ref{ablaStudy}.

Despite simplicity, such a minor modification brings consistent gains over the original Single-Network under same settings. With our discovered augmentation combinations $T_{JOCO}$ and $T_{JCCM}$, the boosted Single-Network can surpass the original Dual-Network evidently. Our ablation studies reveal that the performance gain is non-trivial. We conjecture that regularizing multiple hard augmentations with a shared easy augmentation can be regarded as enforcing consistency among advanced augmentations as well, which inherits the concept of training positive-negative paired samples in contrastive learning \cite{chen2020simple, he2020momentum, chen2021exploring} and SSL-related variations \cite{li2021comatch, yang2022class, wu2023chmatch}.

\subsubsection{Effectiveness Verification}\label{EffVer}
As shown in Tab.~\ref{tabMultiAugsAS} and Fig.~\ref{MultiAugsAS}, we also experimentally verified the major advantage of the multi-path augmentations (dubbed as MultiAugs) strategy. Here, we have two variables of MultiAugs: the number of paths and the category of augmentations. For acquisition of an optimal augmentations set, similarly, we continue with the rules summarized in the previous section, and elect 12 different MultiAugs schemes for illustrating. We can witness the effectiveness of MultiAugs from two aspects: (1) \textit{It can inherit and even expand the synergistic effects between different augmentations.} (2) \textit{It can alleviate the defects caused by excessive stacking of augmentations.}

\vspace{0.2cm}
\makeatletter\def\@captype{table}\makeatother
\hspace*{-0.3cm}\begin{minipage}{.23\textwidth}\footnotesize  
	\centering
	\caption{Best mAP results of different MultiAugs tests.}
	\vspace{-5pt}
	\label{tabMultiAugsAS}
	\setlength{\tabcolsep}{3pt}
	\begin{tabular}{c|c|c}
	\Xhline{1.2pt}
	Id & MultiAugs & mAP \\
	\Xhline{1.2pt}
	$m_1$ & $T_{JC},T_{CM}$ & 43.0 \\
	$m_2$ & $T_{JO},T_{CO}$ & 43.4 \\
	$m_3$ & $T_{JC},T_{JO}$ & 43.1 \\
	\hline
	$m_4$ & $T_{JCCM},T_{CO}$ & 43.7 \\
	$m_5$ & $T_{JOCO},T_{CM}$ & 43.1 \\
	$m_6$ & $T_{JOCO},T_{JC}$ & 44.2 \\
	$m_7$ & $T_{JCCM},T_{JO}$ & 42.9 \\
	\hline
	$m_8$ & \makecell{$T_{JC},T_{JO},$\\$T_{CO},T_{CM}$} & 43.6 \\
	\hline
	$m_9$ & $T_{JCCM}$ (twice)  & 43.6 \\
	\hline
	$m_{10}$ & $T_{JOCO}$ (twice) & 44.0 \\
	\hline
	\rowcolor{gray!20}
	$m_{11}$ & \Gape[0pt][2pt]{\makecell{$T_{JOCO},$\\$T_{JCCM}$}} & 44.9 \\
	\hline
	\rowcolor{gray!20}
	$m_{12}$ & \Gape[0pt][2pt]{\makecell{$T_{JOCO},T_{JC},$\\$T_{JCCM},T_{JO}$}} & {\bf 45.5} \\
	\Xhline{1.2pt}
	\end{tabular}
\end{minipage}
\makeatletter\def\@captype{figure}\makeatother
\begin{minipage}{.25\textwidth} 
	\centering
	\includegraphics[width=\linewidth]{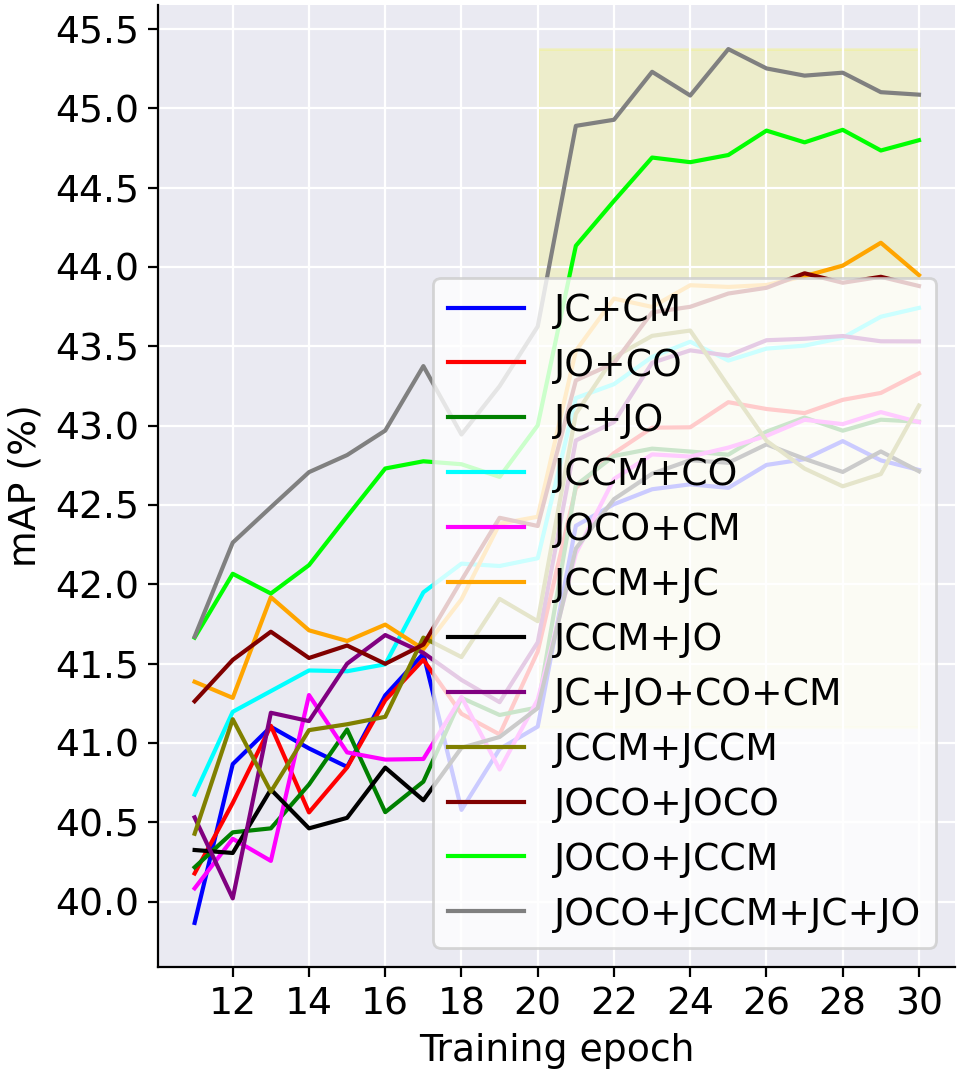} \\
	\vspace{-5pt}
	\caption{The corresponding convergence curves of different tests in Tab.~\ref{tabMultiAugsAS}.}
	\label{MultiAugsAS}
\end{minipage}
\vspace{0.1cm}

Specifically, by comparing $c_1$-$m_1$, $c_2$-$m_2$ and $c_7$-$m_3$, we find that MultiAugs provides comparable performance with the same augmentations. By comparing $c_{10}$-$m_6$, $c_{11}$-$m_7$, $c_{12}$-$m_4$ and $c_{13}$-$m_5$, we can observe the sustained large gains brought by MultiAugs. The scheme $m_8$ does not obtain a better result than $m_{11}$ showing that new augmentations $T_{JOCO}$ and $T_{JCCM}$ have their essential properties. Schemes $m_9$ and $m_{10}$ which utilize a single augmentation twice are also inferior to $m_{11}$ showing the cooperativity of using multi-path distinct augmentations. Moreover, the optimal scheme $m_{12}$ further unleashes capabilities of four advanced augmentations. In order to balance performance and time consumption, we do not add more augmentation paths.

\subsection{Analysis of Superior Augmentations}\label{studyAnaly}

\begin{figure}[t]
	\centering
	\subfloat[{\footnotesize ~}]{\includegraphics[width=0.481\columnwidth]{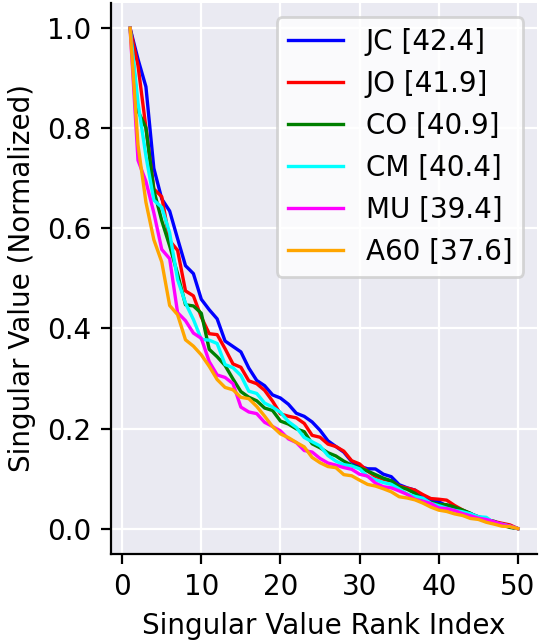}
	\label{plotSVD}}
	\subfloat[{\footnotesize ~}]{\includegraphics[width=0.481\columnwidth]{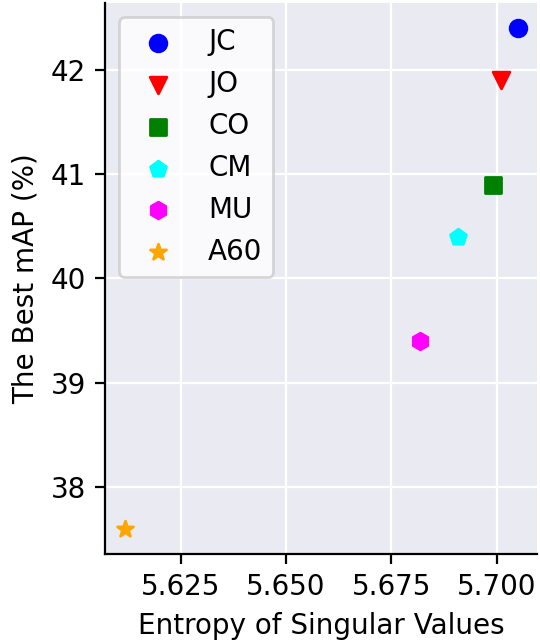}
	\label{plotSVE}}
	\vspace{-5pt}
	\caption{(a) The top-50 singular values of different augmentations. (b) The mAP \textit{vs.} Entropy of different augmentations.}
	\vspace{-10pt}
	\label{plotSVDSVE}
\end{figure}

In this part, we tentatively analyze why employing a superior augmentation to strongly augment the unlabeled data can improve model performance. Different from UDA \cite{xie2020unsupervised} using the improved connectivity of constructed graphs to explain, we start from the perspective of shaped feature space of unlabeled data based on the singular value spectrum, which is widely considered to be related to the model transferability and generalization \cite{chen2019transferability, xue2022investigating}. Specifically, we perform singular value decomposition (SVD) on features $\mathbf{F}\!\in\!\mathbb{R}^{N \times D}$\footnote{We denote $N$ as the number of samples and $D$ as feature dimensions (\textit{a.k.a}, $D \le N$).} extracted by various trained models with different strong augmentations on one dataset: $\mathbf{F}\!=\!\mathbf{U}\mathbf{\Sigma}\mathbf{V}^{T}$, where $\mathbf{U}$ and $\mathbf{V}$ is the left and right singular vector matrices, respectively, and $\mathbf{\Sigma}$ denotes the diagonal singular value matrix $\{\sigma_1, \sigma_2, ... , \sigma_D\}$. Then, we plot calculated top-50 singular values in Fig.~\ref{plotSVD}. In our SVD analysis, the backbone of all six models is ResNet18. The 512-D features of 6,352 samples in COCO \textit{val-set} are extracted. To further measure the flatness of the singular value distribution, we calculate the entropy of normalized singular values $\mathsf{H}_{nsv}$:
\begin{equation}  
	\mathsf{H}_{nsv} = -\sum\nolimits^D_{i=1}\frac{\sigma_i}{\sum\nolimits^D_{j=1}\sigma_j}\log{\frac{\sigma_i}{\sum\nolimits^D_{j=1}\sigma_j}}. ~
	\label{entropySingularValues}
\end{equation}
Usually, a larger $\mathsf{H}_{nsv}$ indicates that the feature space captures more structure in the data and thus spans more dimensions due to more discriminated representations learned. As shown in Fig.~\ref{plotSVE}, the model performance is \textit{positively correlated} with the $\mathsf{H}_{nsv}$ value. Therefore, a superior augmentation facilitates better model generalization to unseen test sets.

\section{Overall Framework}\label{methodMAs}

\begin{figure*}[t]
	\centering
	\subfloat[{\footnotesize MultiAugs with Single-Network}]{\includegraphics[width=0.353\textwidth]{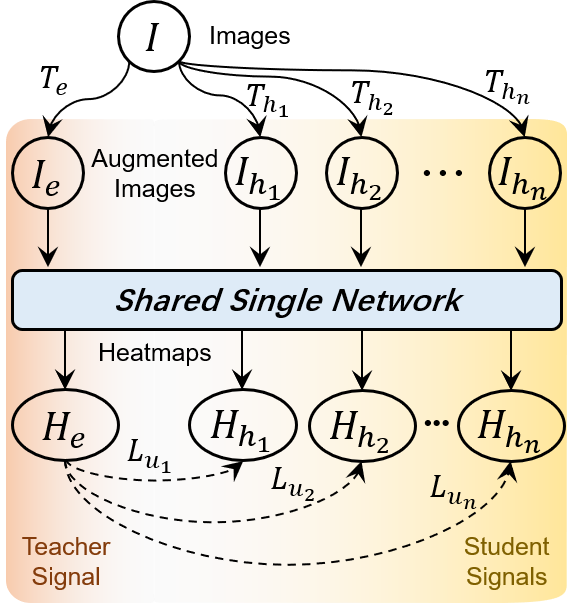}
	\label{NewA}}
	\subfloat[{\footnotesize MultiAugs with Dual-Network}]{\includegraphics[width=0.636\textwidth]{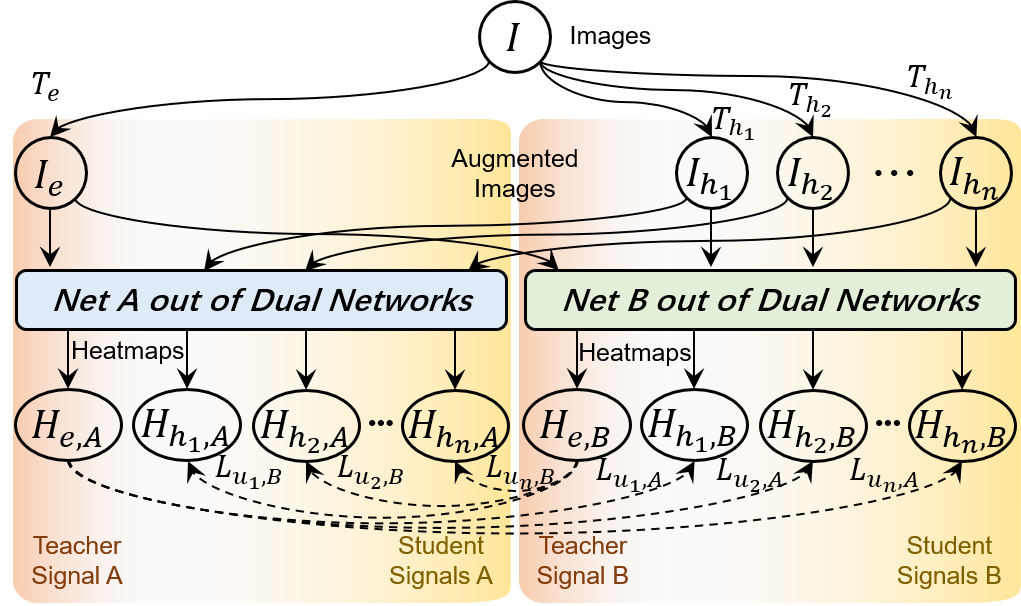}
	\label{NewB}}
	\caption{Our proposed MultiAugs, which can utilize multiple hard augmentations and facilitate multi-path consistency training.}
	\vspace{-10pt}
	\label{frameworkNew}
\end{figure*}

We exploit unlabeled images by applying multiple augmentations with integrating two key techniques introduced in Sec.~\ref{studyAugs} and Sec.~\ref{studyMulti}. Firstly, assuming that we have obtained an optimal augmentation set $\mathcal{\widehat{T}}^n\!=\!\{\widehat{T}_{h_i}|^n_{i=1}\}$, where $\widehat{T}_{h_i}$ may be an old augmentation or a novel discovered one. Then, we present our overall training framework in Fig.~\ref{frameworkNew} based on either Single-Network or Dual-Network.

\subsubsection{MultiAugs (Single-Network)}
This is a consistency-based approach. We only need to maintain a single model as in Fig.~\ref{NewA} during training. For each input batch with equal number of labeled images $\mathbf{I}^l$ and unlabeled images $\mathbf{I}^u$, we calculate the supervised loss with the ground-truth heatmaps as in Eq.~\ref{losssup}, and the multiple unsupervised losses as in Eq.~\ref{lossesunsup}, respectively. The final loss is obtained by adding the two loss functions $\mathcal{L}=\mathcal{L}_s + \lambda \mathcal{L}^*_u$ with $\lambda=1$. Note that we only pass the gradient back through $n$ hard augmentations $\mathcal{\widehat{I}}^n$ for generating teacher signals to avoid collapsing. Based on this boosted Single-Network, we complete all ablation experiments by changing the augmentation categories and quantities in $\mathcal{\widehat{T}}^n$ for controlling the unsupervised loss factor $\mathcal{L}^*_u$.

\subsubsection{MultiAugs (Dual-Network)}
As shown in Fig.~\ref{NewB}, this framework learns two identical yet independent networks with each similar to the Single-Network. For one input batch in every step, each of the two networks serves as both a teacher and a student. They both are fed by easy and hard augmentations of unlabeled images $\mathbf{I}^u$ when they produce teacher signals and student signals, respectively. Assuming we have two networks $f_A$ and $f_B$, and also the augmented easy images $\mathbf{I}^u_e$ using $T_e$ and $n$-path hard images $\{\mathbf{I}^u_{h_1}, \mathbf{I}^u_{h_2}, ..., \mathbf{I}^u_{h_n}\}$ using $\mathcal{\widehat{T}}^n$, we first predict the following four types of heatmaps:
\begin{equation}  
\begin{aligned}
	&\mathbf{H}_{e,A}=T_{A30 \rightarrow A60}(f_A(\mathbf{I}^u_e)), \\~
	&\mathcal{H}_{A}=\{\mathbf{H}_{h_i,A}|^n_{i=1}, \mathbf{H}_{h_i,A}=f_A(\mathbf{I}^u_{h_i})\}, \\~
	&\mathbf{H}_{e,B}=T_{A30 \rightarrow A60}(f_B(\mathbf{I}^u_e)), \\~
	&\mathcal{H}_{B}=\{\mathbf{H}_{h_i,B}|^n_{i=1}, \mathbf{H}_{h_i,B}=f_B(\mathbf{I}^u_{h_i})\},
	\label{heatDualnet}
\end{aligned}
\end{equation}
where $T_{A30 \rightarrow A60}$ is a pre-generated affine transformation. Based on above heatmaps, we calculate two unsupervised losses for training two networks as follows:
\begin{equation}  
\begin{aligned}
	&\mathcal{L}^*_{u,A} = \mathbb{E}_{\mathbf{H}_{h_i,A}\in\mathcal{H}_A}\sum\nolimits^n_{i=1} || \mathbf{H}_{e,B} - \mathbf{H}_{h_i,A} ||^2, \\~
	&\mathcal{L}^*_{u,B} = \mathbb{E}_{\mathbf{H}_{h_i,B}\in\mathcal{H}_B}\sum\nolimits^n_{i=1} || \mathbf{H}_{e,A} - \mathbf{H}_{h_i,B} ||^2, ~
	\label{lossesUnsupDualnet}
\end{aligned}
\end{equation}
where we swap positions of teacher signals $\mathbf{H}_{e,A}$ and $\mathbf{H}_{e,B}$ for realizing the cross training of networks $f_B$ and $f_A$. Following \cite{xie2021empirical}, we report the average accuracy of the final two well-trained and performance-approached models. Besides, $f_A$ and $f_B$ can have different structures as in \cite{xie2021empirical, huang2023semi}, where the large one often helps to distill a better small model, but not vice versa. We thus do not explore this consensus anymore.

\section{Experiments}\label{exps}

\subsection{Datasets and Setups}

\subsubsection{Benchmarks}
Our experiments are mainly conducted on several HPE datasets such as COCO \cite{lin2014microsoft}, MPII \cite{andriluka20142d}, AI-Challenger \cite{wu2019large}, CEPDOF \cite{duan2020rapid}, WEPDTOF \cite{tezcan2022wepdtof} and COCO-WholeBody Hands \cite{jin2020whole}. The first three benchmarks are mainly used to study 2D human keypoint under conventional undistorted cameras. Specifically, the dataset COCO \cite{lin2014microsoft} has 4 subsets: {\it train-set} (118K images), {\it val-set} (5K images), {\it test-dev} and {\it test-challenge}. It is a popular large-scale benchmark for human pose estimation, which contains over 150K annotated people with each represented by 17 keypoints. In addition, there are 123K wild unlabeled images ({\it wild-set}). We selected the first 1K, 5K and 10K samples from {\it train-set} as the labeled set. In some experiments, unlabeled data came from the remaining images of {\it train-set}. In other experiments, we used the whole {\it train-set} as the labeled dataset and {\it wild-set} as the unlabeled dataset. The metric of mAP (Average AP over 10 OKS thresholds) is reported. The dataset MPII \cite{andriluka20142d} has 25K images and 40K person instances with 16 keypoints. The dataset AI-Challenger (AIC)  \cite{wu2019large} {\it train-set} has 210K images and 370K person instances with 14 keypoints. We use MPII as the labeled set, AIC as the unlabeled set. The metric of PCKh\@0.5 is reported.

To further verify the effectiveness of our method, we conducted experiments on an indoor overhead fisheye human keypoint dataset WEPDTOF-Pose which is based on CEPDOF \cite{duan2020rapid} and WEPDTOF \cite{tezcan2022wepdtof}. Following SSPCM \cite{huang2023semi}, we used the complete WEPDTOF-Pose \textit{train-set} (4,688 person instances) as the labeled dataset, and CEPDOF \cite{duan2020rapid} with 11,878 person instances as the unlabeled dataset for experiment. The WEPDTOF-Pose \textit{test-set} (1,179 person instances) is used as the evaluation set. The metric of mAP \cite{lin2014microsoft} is reported for comparing. More details of WEPDTOF-Pose dataset can be found in \cite{huang2023semi}. It should be noted that SSPCM does not open source the BKFisheye dataset included in WEPDTOF-Pose, so we cannot conduct corresponding experimental comparisons involving the BKFisheye with it. Besides, to further verify the wide applicability and superiority of MultiAugs, we also conducted experiments on the human hand keypoint detection task. This is very similar to the human body pose estimation. Specifically, we selected relevant human hand images and corresponding keypoint annotations from the COCO-WholeBody dataset \cite{jin2020whole} as the annotated dataset (where the number of hand keypoints is 21), and obtained a train-set and a val-set containing approximately 76K and 3.8K samples, respectively. In addition, we used the BPJDet detector \cite{zhou2024bpjdet} to extract approximately 118K samples from COCO wild-set as the unlabeled dataset. This hand-oriented dataset can serve as a supplement to other trials of the whole body.

\begin{table}[!]  
	\centering
	\caption{Detailed description of all seven training settings.}
	\vspace{-5pt}
	\label{tabSettings}
	\setlength{\tabcolsep}{2pt}
	\begin{tabular}{c|c|c|c|c}
	\Xhline{1.2pt}
	\makecell{Setting \\Name} & \makecell{Labeled \\Dataset} & \makecell{Unlabeled \\Dataset} & \makecell{Evaluation \\Dataset} & Domain \\
	\hline
	{\bf S1} & \makecell{COCO train-set\\ (front $x\%$)} & \makecell{COCO train-set\\ (left $1-x\%$)} & COCO val-set & body + normal \\
	\hline
	{\bf S2} & COCO train-set & COCO wild-set & COCO val-set & body + normal \\
	\hline
	{\bf S3}  & COCO train-set & COCO wild-set & COCO test-set & body + normal \\
	\hline
	{\bf S4}  & MPII train-set & AIC train-set & MPII val-set & body + normal \\
	\hline
	{\bf S5}  & MPII train-set & AIC train-set & MPII test-set & body + normal\\
	\hline
	{\bf S6}  & \makecell{WEPDTOF-\\Pose train-set} & CEPDOF & \makecell{WEPDTOF-\\Pose test-set}  & body + fisheye \\
	\hline
	{\bf S7}  & \makecell{COCO Whole-\\Body train-set}  & COCO wild-set & \makecell{COCO Whole-\\Body val-set}  & hand + normal \\
	\Xhline{1.2pt}
	\end{tabular}
	\vspace{-10pt}
\end{table}
 
\subsubsection{Training Settings}
First, we formulate five SSL training experimental settings (including {\bf S1}, {\bf S2}, {\bf S3}, {\bf S4} and {\bf S5}) based on the arrangements in various previous SSHPE methods \cite{xie2021empirical, huang2023semi, yu2024denoising}. These setting mainly involve datasets COCO, MPII and AIC containing conventional human body keypoints. Then, we employ the sixth setting ({\bf S6}) based on the dataset WEPDTOF-Pose, which mainly includes the images of human body keypoints with obvious distortion under the fisheye camera. Finally, we newly add the seventh setting ({\bf S7}) focusing on the human hand keypoints detection task based on the dataset COCO-WholeBody Hands. The last two settings are domains that are different from mainstream 2D HPE tasks but have essential similarities, which is very suitable for us to demonstrate the universality of MultiAugs. We summarize the specifics of these settings in Tab.~\ref{tabSettings}.

\subsubsection{Implementation Details}
For a fair comparison with prior works, we use SimpleBaseline \cite{xiao2018simple} to estimate heatmaps and ResNet \cite{he2016deep} and HRNet \cite{sun2019deep} as backbones. The input image size is set to $256\times192$. We adopt the PyTorch 1.30 and 4 A100 GPUs with each batch size as 32 for training. We use the Adam optimizer to train these models. The initial learning rate is 1e-3. When training on COCO with 10K labeled data, it decreases to 1e-4 and 1e-5 at epochs 70 and 90, respectively, with a total of 100 epochs. When using 1K or 5K labeled data, total epochs are reduced to 30 or 70, respectively. When training on the complete COCO or MPII+AIC, it drops to 1e-4 and 1e-5 at epochs 220 and 260, respectively, with a total of 300 epochs. When training on WEPDTOF-Pose, it decreases to 1e-4 and 1e-5 at 140 epochs and 180 epochs, respectively, with a total of 200 epochs. When training on COCO-WholeBody, we repeat the similar setting as the COCO dataset.

For data augmentation, we keep the easy augmentation $T_e$ as $T_{A30}$. In Sec.~\ref{SelAugCom} and Sec.~\ref{EffVer}, we have presented details of finding two novel superior hard augmentations (\eg, $T_{JOCO}$ and $T_{JCCM}$) and recommending the optimal multi-path augmentation set $\mathcal{\widehat{T}}^4$ (please see the scheme $m_{12}$ in Tab. \ref{tabMultiAugsAS}), repsectively. Therefore, the hard augmentations set $\mathcal{\widehat{T}}^4\!=\!\{T_{JOCO}, T_{JCCM}, T_{JC}, T_{JO}\}$ is chosen for settings {\bf S1}, {\bf S2}, {\bf S3}, {\bf S6} and {\bf S7}. While, the less optimal augmentations set $\mathcal{\widehat{T}}^2\!=\!\{T_{JOCO}, T_{JCCM}\}$ is chosen for settings {\bf S4} and {\bf S5} to balance the performance and training time.

\begin{table}[!]  
	\centering
	\caption{Hyper-parameters details of basic augmentations.}
	\vspace{-5pt}
	\label{tabParams}
	\setlength{\tabcolsep}{4pt}
	\begin{tabular}{l|c|c}
	\Xhline{1.2pt}
	Aug. & Type & Description \\
	\hline
	$T_{A30}$ & easy & \makecell{random scale within range $[0.75, 1.25]$, \\random rotation within range $(-30^\circ, 30^\circ)$.} \\
	\hline
	$T_{A60}$  & hard & \makecell{random scale within range $[0.75, 1.25]$, \\random rotation within range $(-60^\circ, 60^\circ)$.} \\
	\hline
	$T_{CO}$  & hard & \makecell{random generation \textbf{5} zero-valued size $20 \times 20$ patches.} \\
	\hline
	$T_{CM}$  & hard & \makecell{random cropping \textbf{2} $20 \times 20$ patches from other images.} \\
	\hline
	$T_{JC}$  & hard & \makecell{random generation \textbf{5} zero value \\patches with 
		size $20 \times 20$ around predicted joints.} \\
	\hline
	$T_{JO}$  & hard & \makecell{random cropping \textbf{2} patches from \\other images with 
		size $20 \times 20$ around predicted joints.} \\
	\Xhline{1.2pt}
	\end{tabular}
	\vspace{-10pt}
\end{table}

\subsubsection{Hyper-parameters of Basic Augmentations}
The hyper-parameters involved in each augmentation are indeed important. In order to make a fair comparison, we list the parameters of the basic augmentations used in this paper in Tab.~\ref{tabParams}, so that readers can quickly and clearly know these details. Particularly, each basic augmentation we selected is derived from various compared methods without additional fine-tuning. For example, the parameters of Joint Cutout are the same as those in PoseDual \cite{xie2021empirical} which used JC5, and the parameters of Joint Cut-Occlude are the same as those in SSPCM \cite{huang2023semi} which used JO2. Please refer Fig.~\ref{AugsPlus} for some visualization results after applying augmentations.

\subsection{Performance Comparison}

We mainly compare with representative SSHPE methods including PoseDual \cite{xie2021empirical}, SSPCM \cite{huang2023semi} and Pseudo-HMs \cite{yu2024denoising} under various settings. Note that Pseudo-HMs does not follow the same setup as PoseDual \cite{xie2021empirical} and SSPCM \cite{huang2023semi}, and its code has not been released. We have tried our best to list partial comparable data in some tables to maintain completeness.

\textbf{S1}: Firstly, we conduct experiments on the COCO {\it train-set} with 1K, 5K and 10K labeled data, and evaluate on the {\it val-set}. As shown in Tab.~\ref{tab1}, our method brings substantial improvements under the same setting. For example, when using a Single-Network, our method exceeds both PoseCons and PoseDual significantly, and is close to the SSPCM based on three networks. When using a Dual-Network, our method exceeds previous SOTA results by {\bf 2.8} mAP, {\bf 1.3} mAP, and {\bf 1.1} mAP under 1K, 5K and 10K settings, respectively. Note that our method can bring greater gains with less labeled data (\eg, 1K images), which further explains its efficiency and superiority. In addition, we consider that reporting the comparison results based on the more common ResNet50 is more convincing than ResNet18. Therefore, we replace the backbone in Tab.~\ref{tab1} with ResNet50 and re-conduct the comparative experiments. The results are shown in Tab.~\ref{tab7}. Similar to using ResNet18, our method can still achieve a clear advantage. When using a single network, our method outperforms PoseCons and Posedual, while being comparable to SSPCM. And our dual-network based approach achieves significant advantages.

\begin{table}[!]  
	\centering
	\caption{AP of different methods on COCO {\it val-set} when different numbers of labels are used. The backbone of all methods is {\bf ResNet18}. The best results are in bold.}
	\vspace{-5pt}
	\label{tab1}
	\setlength{\tabcolsep}{7pt}
	\begin{tabular}{l|c|cccc}
	\Xhline{1.2pt}
	\multirow{2}{*}{Method} & \multirow{2}{*}{Nets} & \multicolumn{4}{c}{Labeled Samples} \\
	\cline{3-6}
	~ & ~ & 1K & 5K & 10K & All \\
	\hline
	Supervised \cite{xiao2018simple} & 1 & 31.5 & 46.4 & 51.1 & 67.1 \\
	\hline
	PseudoPose \cite{xie2021empirical} & 2 & 37.2 & 50.9 & 56.0 & --- \\
	DataDistill \cite{radosavovic2018data} & 2 & 37.6 & 51.6 & 56.6 & --- \\
	PoseCons \cite{xie2021empirical} & 1 & 42.1 & 52.3 & 57.3 & --- \\
	PoseDual \cite{xie2021empirical} & 2 & 44.6 & 55.6 & 59.6 & --- \\
	SSPCM \cite{huang2023semi} & 3 & 46.9 & 57.5 & 60.7 & --- \\
	Pseudo-HMs \cite{yu2024denoising} & 2 & 47.6 & --- & --- & --- \\
	\hline
	Ours (Single) & 1 & 45.5 & 56.2 & 59.9 & --- \\
	Ours (Dual) & 2 & {\bf 49.7} & {\bf 58.8} & {\bf 61.8} & --- \\
	\Xhline{1.2pt}
	\end{tabular}
	\vspace{-5pt} 
\end{table}

\begin{table}[!] 
	\centering
	\caption{AP of different methods on COCO {\it val-set} when different numbers of labels are used. The backbone of all methods is {\bf ResNet50}. The best results are in bold.}
	\vspace{-5pt}
	\label{tab7}
	\setlength{\tabcolsep}{7pt}
	\begin{tabular}{l|c|cccc}
	\Xhline{1.2pt}
	\multirow{2}{*}{Method} & \multirow{2}{*}{Nets} & \multicolumn{4}{c}{Labeled Samples} \\
	\cline{3-6}
	~ & ~ & 1K & 5K & 10K & All \\
	\hline
	Supervised \cite{xiao2018simple} & 1 & 34.8 & 50.6 & 56.4 & 70.9 \\
	\hline
	PoseCons \cite{xie2021empirical} & 1 &  43.1 & 57.2 & 61.8 & --- \\
	PoseDual \cite{xie2021empirical} & 2 & 48.2 & 61.1 & 65.0 & --- \\
	SSPCM \cite{huang2023semi} & 3 & 49.8 & 61.8 & 65.5 & --- \\
	\hline
	Ours (Single) & 1 & 49.3 & 61.4 & 65.2 & --- \\
	Ours (Dual) & 2 & {\bf 51.7} & {\bf 62.9} & {\bf 66.3} & --- \\
	\Xhline{1.2pt}
	\end{tabular}
	\vspace{-10pt}
\end{table}

\textbf{S2}: Then, we conduct larger scale SSHPE experiments on the complete COCO dataset by using {\it train-set} as the labeled dataset and {\it wild-set} as the unlabeled dataset. As shown in Tab.~\ref{tab2}, regardless of using any backbone, our method can always improve all supervised baseline results, and bring more gains than two compared SSHPE methods \cite{xie2021empirical} and \cite{huang2023semi} with using dual networks. When using a single network, our method is still superior to PoseDual \cite{xie2021empirical}, and fairly close to SSPCM \cite{huang2023semi} based on triple networks.

\textbf{S3}: We also report results using HRNet-w48 on the COCO {\it test-dev} in Tab.~\ref{tab3}. The upper, middle and lower panels show CNN-based supervised, Transformers-based supervised, and SSHPE methods, respectively. Our method can slightly outperform the PoseDual but fall behind the best SSPCM. We attribute it to our fewer training epochs (300 vs. 400) and less parameters (2 Nets vs. 3 Nets) which may lead to weaker generalization. Besides, we can observe that MultiAugs outperforms some burdensome transformer-based methods \cite{yang2021transpose, li2021tokenpose, yuan2021hrformer}, which reveals the significance of rational utilization of unlabeled data and advanced SSL techniques.

\begin{table}[!] 
	\centering
	\caption{Results on COCO {\it val-set} with labeled {\it train-set} and unlabeled {\it wild-set} for SSL training. The best results are in bold.}
	\vspace{-5pt}
	\label{tab2}
	\setlength{\tabcolsep}{4pt}
	\begin{tabular}{l|c|c|cccc}
	\Xhline{1.2pt}
	Method & Backbone & Nets & AP & AP$_{.5}$ & AR & AR$_{.5}$ \\
	\hline
	Supervised \cite{xiao2018simple} & ResNet50 & Single & 70.9 & 91.4 & 74.2 & 92.3 \\
	PoseDual \cite{xie2021empirical} & ResNet50 & Dual & 73.9 & 92.5 & 77.0 & 93.5 \\
	SSPCM \cite{huang2023semi} & ResNet50 & Triple & 74.2 & 92.7 & 77.2 & 93.8 \\
	Pseudo-HMs \cite{yu2024denoising} & ResNet50 & Dual & 74.1 & --- & --- & --- \\
	Ours (Single) & ResNet50 & Single & 74.4 & {\bf 93.6} & 77.4 & {\bf 94.0} \\ 
	Ours (Dual) & ResNet50 & Dual & {\bf 74.6} & 93.5 & {\bf 77.6} & {\bf 94.0}  \\ 
	\hline
	Supervised \cite{xiao2018simple} & ResNet101 & Single & 72.5 & 92.5 & 75.6 & 93.1 \\
	PoseDual \cite{xie2021empirical} & ResNet101 & Dual & 75.3 & 93.6 & 78.2 & 94.1 \\
	SSPCM \cite{huang2023semi} & ResNet101 & Triple & 75.5 & {\bf 93.8} & 78.4 & 94.2 \\
	Pseudo-HMs \cite{yu2024denoising} & ResNet101 & Dual & 75.7 & --- & --- & --- \\
	Ours (Single) & ResNet101 & Single & 75.8 & 93.5 & 78.8 & 94.4 \\ 
	Ours (Dual) & ResNet101 & Dual & {\bf 76.4} & 93.6 & {\bf 79.3} & {\bf 94.7} \\ 
	\hline
	Supervised \cite{xiao2018simple} & HRNet-w48 & Single & 77.2 & 93.5 & 79.9 & 94.1 \\
	PoseDual \cite{xie2021empirical} & HRNet-w48 & Dual & 79.2 & 94.6 & 81.7 & 95.1 \\
	SSPCM \cite{huang2023semi} & HRNet-w48 & Triple & 79.4 & {\bf 94.8} & 81.9 & {\bf 95.2} \\
	Pseudo-HMs \cite{yu2024denoising} & HRNet-w48 & Dual & 79.4 & --- & --- & --- \\
	Ours (Single) & HRNet-w48 & Single & 79.3 & 94.6 & 81.9 & 95.1 \\ 
	Ours (Dual) & HRNet-w48 & Dual & {\bf 79.5} & 94.6 & {\bf 82.1} & {\bf 95.2} \\ 
	\Xhline{1.2pt}
	\end{tabular}
	\vspace{-5pt}
\end{table}

\begin{table}[t]  
	\centering
	\caption{Comparison to SOTA methods on the COCO {\it test-dev}. Person detection results are provided by SimpleBaseline \cite{xiao2018simple} and flipping strategy is used. The best and second best results are in bold and underlined, respectively.}
	\vspace{-5pt}
	\label{tab3}
	\setlength{\tabcolsep}{1.2pt}
	\begin{tabular}{l|c|c|c|c|cc}
	\Xhline{1.2pt}
	Method & Backbone & Input Size & Gflops & Params & AP & AR \\
	\hline
	 SimpleBaseline \cite{xiao2018simple} & ResNet50 & 256$\times$192 & 8.9 & 34.0 & 70.2 & 75.8 \\
	HRNet \cite{sun2019deep} & HRNet-w48 & 384$\times$288 & 32.9 & 63.6 & 75.5 & 80.5 \\
	MSPN \cite{li2019rethinking} & ResNet50 & 384$\times$288 & 58.7 & 71.9 & 76.1 & 81.6 \\
	DARK \cite{zhang2020distribution} & HRNet-w48 & 384$\times$288 & 32.9 & 63.6 & 76.2 & 81.1 \\  
	UDP \cite{huang2020devil} & HRNet-w48 & 384$\times$288 & 33.0 & 63.8 & 76.5 & 81.6 \\
	\hline
	TransPose-H-A6 \cite{yang2021transpose} & HRNet-w48 & 256$\times$192 & 21.8 & 17.5 & 75.0 & --- \\
	TokenPose-L/D24 \cite{li2021tokenpose} & HRNet-w48 & 384$\times$288 & 22.1 & 29.8 & 75.9 & 80.8 \\
	HRFormer \cite{yuan2021hrformer} & HRFormer-B & 384$\times$288 & 26.8 & 43.2 & 76.2 & 81.2 \\
	ViTPose \cite{xu2022vitpose} & ViT-Large & 256$\times$192 & 59.8 & 307.0 & 77.3 & 82.4 \\
	\hline 
	DUAL (+HRNet) \cite{xie2021empirical} & HRNet-w48 & 384$\times$288 & 65.8 & 127.2 & 76.7 & 81.8 \\
	DUAL (+DARK) \cite{xie2021empirical} & HRNet-w48 & 384$\times$288 & 65.8 & 127.2 & 77.2 & 82.2 \\
	SSPCM (+DARK) \cite{huang2023semi} & HRNet-w48 & 384$\times$288 & 98.7 & 190.8 & {\bf 77.5} & {\bf 82.4} \\
	Ours (Dual) (+HRNet) & HRNet-w48 & 384$\times$288 & 65.8 & 127.2 & 76.8 & 81.8 \\ 
	Ours (Dual) (+DARK) & HRNet-w48 & 384$\times$288 & 65.8 & 127.2 & \underline{77.4} & \underline{82.3} \\ 
	\Xhline{1.2pt}
	\end{tabular}
	\vspace{-10pt}
\end{table}

\begin{table}[t]  
	\begin{center}
	\caption{Results on {\it val-set} of MPII. HRNet is trained only on MPII {\it train-set}. The ``*'' means using extra labeled dataset AIC. The ``+'' means applying the model ensemble. Best results are in bold.}
	\vspace{-5pt}
	\label{tab4}
	\setlength{\tabcolsep}{4pt}
	\begin{tabular}{l|cccccccc}
	\Xhline{1.2pt}
	Method & Hea & Sho & Elb & Wri & Hip & Kne & Ank & Total \\
	\hline
	HRNet \cite{sun2019deep} & 97.0 & 95.7 & 89.4 & 85.6 & 87.7 & 85.8 & 82.0 & 89.5 \\
	HRNet* \cite{sun2019deep} & 97.4 & 96.7 & 92.1 & 88.4 & 90.8 & 88.6 & 85.0 & {\bf 91.7} \\
	\hline
	PoseDual \cite{xie2021empirical} & 97.4 & 96.6 & 91.8 & 87.5 & 89.6 & 87.6 & 83.8 & 91.1 \\
	Ours (Dual) & 97.3 & 96.8 & 91.7 & 87.5 & 90.3& 88.6 & 84.6 & 91.4 \\
	Ours+ (Dual) & 97.3 & 96.8 & 91.9 & 88.1 & 90.6 & 89.2 & 85.0 & {\bf 91.7} \\
	\Xhline{1.2pt}
	\end{tabular}
	\end{center}
	\vspace{-10pt}
\end{table}

\begin{table}[t]  
	\begin{center}
	\caption{Results on the {\it val-set} of MPII dataset. The used backbone is ResNet18. The best results are in bold.}
	\vspace{-5pt}
	\label{tab8}
	\setlength{\tabcolsep}{4pt}
	\begin{tabular}{l|cccccccc}
	\Xhline{1.2pt}
	Method & Hea & Sho & Elb & Wri & Hip & Kne & Ank & Total \\
	\hline
	Supervised \cite{xiao2018simple} & 89.6 & 84.8 & 72.0 & 58.4 & 57.8 & 49.4 & 41.2 & 65.3 \\
	\hline
	PoseCons \cite{xie2021empirical} & 92.7 & 87.6 & 74.5 & 67.9 & 72.3 & 64.2 & 59.4 & 75.2 \\
	PoseDual \cite{xie2021empirical} & 93.3 & 88.4 & 75.0 &  67.3 & 72.6 & 65.3 & 59.7 & 75.6 \\
	SSPCM \cite{huang2023semi} & 93.5 & 90.6 & 80.2 & 71.3 & 75.9 & 68.9 & 62.3	& 78.3 \\
	Ours (Single) & 94.1 & 91.1 & 80.5 & 72.2 & 76.3 & 69.2 & 62.8 & 79.1 \\
	Ours (Dual) & {\bf 94.7} & {\bf 92.4} & {\bf 81.2} & {\bf 73.3} & {\bf 76.8} & {\bf 70.6} & {\bf 63.9} & {\bf 79.7} \\
	\Xhline{1.2pt}
	\end{tabular}
	\end{center}
	\vspace{-10pt}
\end{table}

\begin{table}[!]  
	\begin{center}
	\caption{Comparisons on {\it test-set} of MPII. We use HRNet-w32 as the backbone. The input image size is 256$\times$256. The MPII (with labels) and AIC (without labels) are used for SSL training.}
	\vspace{-5pt}
	\label{tab5}
	\setlength{\tabcolsep}{4pt}
	\begin{tabular}{l|cccccccc}
	\Xhline{1.2pt}
	Method & Hea & Sho & Elb & Wri & Hip & Kne & Ank & Total \\
	\hline
	Newell \etal \cite{newell2016stacked} & 98.2 & 96.3 & 91.2 & 87.1 & 90.1 & 87.4 & 83.6 & 90.9 \\
	Xiao \etal \cite{xiao2018simple} & 98.5 & 96.6 & 91.9 & 87.6 & 91.1 & 88.1 & 84.1 & 91.5 \\
	Ke \etal \cite{ke2018multi} & 98.5 & 96.8 & 92.7 & 88.4 & 90.6 & 89.4 & 86.3 & 92.1 \\
	Sun \etal \cite{sun2019deep} & 98.6 & 96.9 & 92.8 & 89.0 & 91.5 & 89.0 & 85.7 & 92.3 \\
	Zhang \etal \cite{zhang2019human} & 98.6 & 97.0 & 92.8 & 88.8 & 91.7 & 89.8 & 86.6 & 92.5 \\
	\hline
	Xie \etal \cite{xie2021empirical} & 98.7 & 97.3 & 93.7 & 90.2 & 92.0 & 90.3 & 86.5 & 93.0 \\
	Huang \etal \cite{huang2023semi} & 98.7 & 97.5 & 94.0 & {\bf 90.6} & {\bf 92.5} & {\bf 91.1} & 87.1 & 93.3 \\
	Ours (Dual) & {\bf 98.8} & {\bf 97.6} & {\bf 94.1} & 90.3 & 92.4 & {\bf 91.1} & {\bf 87.2} & {\bf 93.4} \\
	\Xhline{1.2pt}
	\end{tabular}
	\end{center}
	\vspace{-10pt}
\end{table}
\textbf{S4}: For the MPII dataset, we allocate its {\it train-set} as the labeled set and whole AIC as the unlabeled set. This is a more realistic setting where labeled and unlabeled images are from different datasets. The Tab.~\ref{tab4} shows results reported on the MPII {\it val-set}. Our method outperforms both the fully supervised HRNet \cite{sun2019deep} and semi-supervised PoseDual \cite{xie2021empirical} by a large margin with using the same backbone. It is worth noting that our semi-supervised MultiAugs with applying the model ensemble can even approach the supervised HRNet with using extra labeled AIC. Moreover, we deem that conclusions drawn from testing on small-scale data may not necessarily be generalized to other datasets. Therefore, we repeat the comparison in setting \textbf{S1} and Tab.~\ref{tab1} by replacing the dataset from COCO into MPII. Specifically, we conduct experiments using the first 1K samples as labeled data and the left 39K samples as unlabeled data in MPII. The validation set of MPII is used to evaluate. The final comparison results are shown in Tab.~\ref{tab8}. Not surprisingly, our method still maintains a clear lead in performance, both in terms of overall accuracy and the specific accuracy of each joint. These experiments once again demonstrate that our method is indeed universally effective and superior across different datasets.

\textbf{S5}: Finally, in Tab.~\ref{tab5}, our results on MPII {\it test-set} surpass all those of previous fully supervised methods and two semi-supervised counterparts PoseDual and SSPCM. This further validates the effectiveness and superiority of our method.

\begin{table}[t]  
	\begin{center}
	\caption{Comparison of our method to SOTA methods on the dataset WEPDTOF-Pose collected by the fisheye camera.}
	\vspace{-5pt}
	\label{tab6}
	\setlength{\tabcolsep}{10pt}
	\begin{tabular}{l|c|cc}
	\Xhline{1.2pt}
	Method & Nets & AP & AR \\
	\hline
	Supervised \cite{xiao2018simple} & 1 & 49.5 & 53.4 \\
	\hline
	PoseCons \cite{xie2021empirical} & 1 & 54.6 & 58.1 \\
	PoseDual \cite{xie2021empirical} & 2 & 55.1 & 59.0 \\
	SSPCM \cite{huang2023semi} & 3 & 55.6 & 60.0 \\
	\hline
	Ours (Single) & 1 & 56.5 & 60.6 \\
	Ours (Dual) & 2 & {\bf 57.1} & {\bf 61.3} \\
	\Xhline{1.2pt}
	\end{tabular}
	\end{center}
	\vspace{-10pt}
\end{table}
\textbf{S6}: For dataset WEPDTOF-Pose, given the particularity of the fisheye data, the random rotation range used in all hard data augmentations is (-90$^\circ$, 90$^\circ$). It means that $T_{A60}$ is changed into $T_{A90}$. As shown in Tab.~\ref{tab6}, our method always achieves the best AP and AR results whether using a single network or a dual network structure, surpassing the previous SOTA method SSPCM using a triple network. These experiments in the fisheye domain once again demonstrate the superiority and universality of our proposed MultiAugs.

\begin{table}[t] 
	\centering
	\caption{AP of different methods on COCO-WholeBody Hands {\it val-set} when different numbers of labels are used.}
	\vspace{-5pt}
	\label{tabS7.1}
	\setlength{\tabcolsep}{8pt}
	\begin{tabular}{l|c|ccc}
	\Xhline{1.2pt}
	\multirow{2}{*}{Method} & \multirow{2}{*}{Nets} & \multicolumn{3}{c}{Labeled Samples} \\
	\cline{3-5}
	~ & ~ & 1K & 5K & 10K \\
	\hline
	Supervised \cite{xiao2018simple} & 1 & 33.4 & 38.9 & 41.9 \\
	\hline
	PoseCons \cite{xie2021empirical} & 1 & 52.1 & 57.4 & 59.4 \\
	PoseDual \cite{xie2021empirical} & 2 & 55.9 & 60.1 & 62.0 \\
	SSPCM \cite{huang2023semi} & 3 & 58.8 & 62.4 & 64.5 \\
	\hline
	Ours (Single) & 1 & 56.3 & 60.9 & 64.1 \\
	Ours (Dual) & 2 & {\bf 62.8} & {\bf 65.5} & {\bf 67.4} \\
	\Xhline{1.2pt}
	\end{tabular}
	\vspace{-10pt}
\end{table}

\begin{table}[!] 
	\centering
	\caption{Results on the COCO-WholeBody Hands {\it val-set} with using unlabeled COCO {\it wild-set} for SSL training.}
	\vspace{-5pt}
	\label{tabS7.2}
	\setlength{\tabcolsep}{8pt}
	\begin{tabular}{l|c|c|cc}
	\Xhline{1.2pt}
	Method & Backbone & Nets & AP & AR \\
	\hline
	Supervised \cite{xiao2018simple} & ResNet50 & Single & 62.1 & 74.9 \\
	PoseDual \cite{xie2021empirical} & ResNet50 & Dual & 65.9 & 78.3 \\
	SSPCM \cite{huang2023semi} & ResNet50 & Triple & 66.3 & 78.8 \\
	Ours (Single) & ResNet50 & Single & 66.8 & 79.2 \\
	Ours (Dual) & ResNet50 & Dual & {\bf 67.4} & {\bf 79.7} \\
	\hline
	Supervised \cite{xiao2018simple} & ResNet101 & Single & 64.5 & 76.8\\
	PoseDual \cite{xie2021empirical} & ResNet101 & Dual & 68.0 & 80.2 \\
	SSPCM \cite{huang2023semi} & ResNet101 & Triple & 69.1 & 81.1 \\
	Ours (Single) & ResNet101 & Single & 69.9 & 81.7 \\
	Ours (Dual) & ResNet101 & Dual & {\bf 71.8} & {\bf 83.3} \\
	\Xhline{1.2pt}
	\end{tabular}
	\vspace{-10pt}
\end{table}

\textbf{S7}: This setting is for training on COCO-WholeBody Hands. Firstly, we still take ResNet18 as the backbone as in the setup \textbf{S1}, and then conduct comparative experiments using the methods SimpleBaseline, PoseCons, PoseDual, SSPCM, and the proposed MutliAugs. The final conclusion is still impressive. As shown in Tab.~\ref{tabS7.1}, when using a single-network structure, our method is still significantly better than the PoseCons using a single network or the PoseDual using two networks. When we use a dual-network structure, our method is significantly better than the SSPCM using a triple-network. These results and the trends are consistent with HPE performance on the whole body. Secondly, we follow the setup \textbf{S2}, and conduct experiments on COCO-WholeBody Hands by using the annotated train-set as the labeled set and the COCO wild-set as the unlabeled set. Without losing generality, we choose to perform various experiments under backbones ResNet50 and ResNet101. The quantitative results on the val-set are summarized in the Tab.~\ref{tabS7.2}. From these results, we can find that our method still has an undoubted advantage, which is consistent with the phenomenon shown in other tests. In a word, the above experimental results show that the advanced augmentation combination and multi-path consistency loss strategy we proposed is indeed sustainable, effective and easy to promote. For example, for keypoint detection tasks, whether it is human body or hand, MultiAugs has reliable transferability and great potential in versatility. We expect that these experiments can further help to demonstrate the core contribution of this paper.

\begin{figure*}[t]
	\centering
	\includegraphics[width=\textwidth]{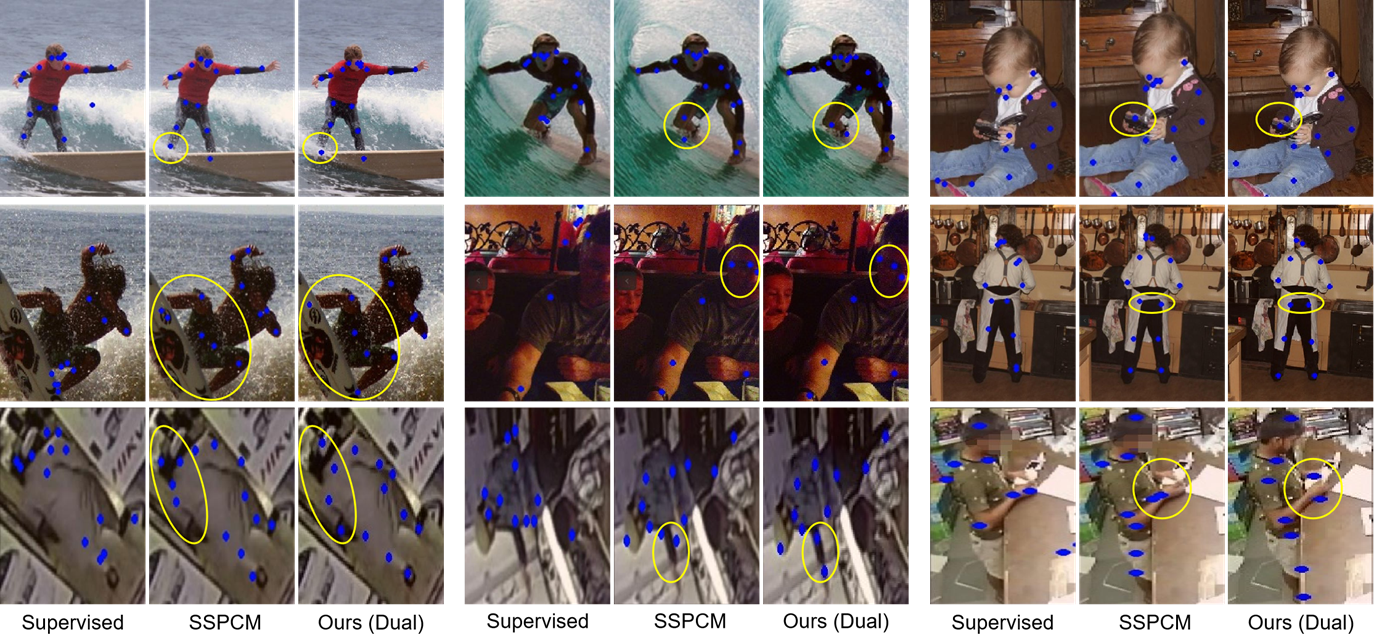}
	\vspace{-15pt}
	\caption{Qualitative results on COCO val-set (the first 6 examples) and WEPDTOF-Pose test-set (the last 3 examples). The predicted results of methods Supervised and SSPCM are directly fetched from the supplementary paper of SSPCM \cite{huang2023semi}.  The details of the comparison between SSPCM and our predictions are highlighted in yellow circles for quick identification.}
	\label{VisCom}
	\vspace{-10pt}
\end{figure*}

\textbf{Qualitative Comparison}:
To make our advantages more intuitively demonstrated, we have presented qualitative visualization comparison results, mainly including the conventional human images from the COCO val-set and the fisheye camera images from the WEPDTOF-Pose dataset. We take use of the backbone ResNet18 for all compared methods to highlight the their performance differences. For models trained on COCO dataset, we use the label set with 10K samples for comparison. As shown in Fig.~\ref{VisCom}, pure supervised learning methods are usually prone to making mistakes or messing up, and other SSHPE methods do not perform well in some occlusion cases or edge keypoint detection. While, our method often obtains more accurate estimation results.

\subsection{Ablation Studies}\label{ablaStudy}

For empirical studies in Fig.~\ref{AugsRanking}, \ref{AugsPlusAS}, \ref{MultiAugsAS} and Tab.~\ref{tabAugsPlusAS}, \ref{tabMultiAugsAS}, we conducted them follow the {\bf S1} using 1K labeled COCO images and the Single-Network framework. Below, with the same setting, we conducted more ablation experiments to further analyze our proposed MultiAugs.

\vspace{0.2cm}
\makeatletter\def\@captype{table}\makeatother
\hspace*{-0.3cm}\begin{minipage}{.23\textwidth}\footnotesize  
	\centering
	\caption{Best mAP results of different augmentations on COCO {\it val-set}.}
	\vspace{-0pt}
	\label{tabOtherAugs}
	\setlength{\tabcolsep}{2.5pt}
	\begin{tabular}{c|c}
	\Xhline{1.2pt}
	Augmentations & mAP \\
	\hline
	\makecell{$T_{RA}$ \cite{cubuk2020randaugment}\\(CVPRW'2020)} & 41.9 \\
	\hline
	\makecell{$T_{TA}$ \cite{muller2021trivialaugment}\\(ICCV'2021)} & 40.2 \\
	\hline
	\makecell{$T_{YOCO}$ \cite{han2022you}+$T_{RA}$\\(ICML'2022)} & 42.5 \\
	\hline
	\makecell{$T_{YOCO}$ \cite{han2022you}+$T_{TA}$\\(ICML'2022)} & 41.6 \\
	\hline
	\makecell{$T_{JC}$ \cite{xie2021empirical}\\(ICCV'2021)} & 42.4 \\
	\hline
	\makecell{$T_{JO}$ \cite{huang2023semi}\\(CVPR'2023)} & 41.9 \\
	\hline
	\makecell{$T_{JCCM}$\\(Ours) } & 42.7 \\
	\hline
	\makecell{$T_{JOCO}$\\(Ours) } & 43.7 \\
	\Xhline{1.2pt}
	\end{tabular}
\end{minipage}
\makeatletter\def\@captype{figure}\makeatother
\begin{minipage}{.25\textwidth} 
	\centering
	\includegraphics[width=\columnwidth]{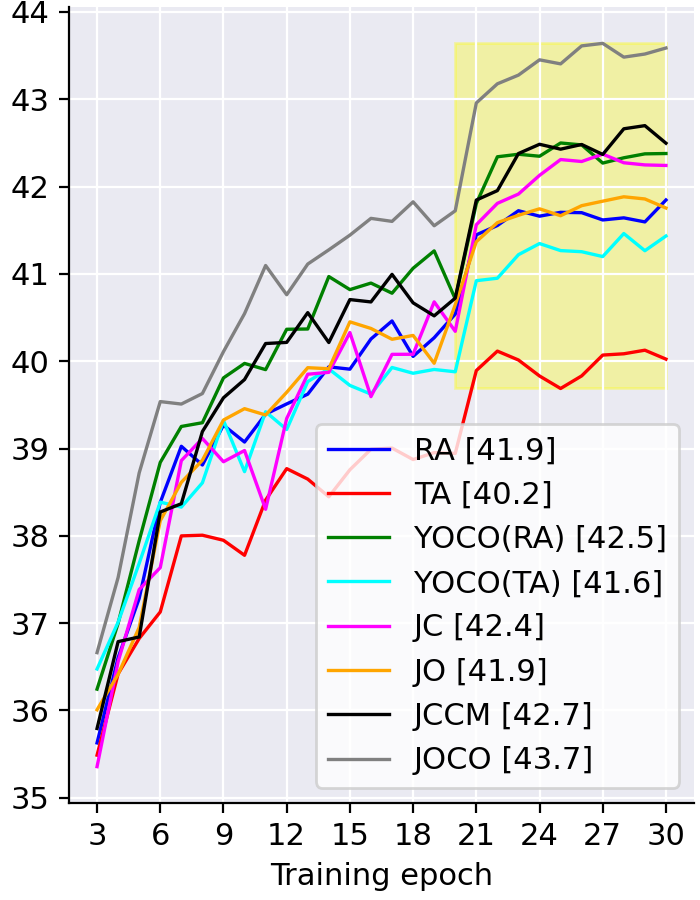}
	\vspace{-18pt}
	\caption{The corresponding convergence curves of different augmentations in Tab.~\ref{tabOtherAugs}.}
	\label{otherAugs}
\end{minipage}
\vspace{0.1cm}

\subsubsection{Comparing to Other Specially Designed Advanced Augmentations}
Three advanced augmentations are selected: RandAugment \cite{cubuk2020randaugment}, TrivialAugment \cite{muller2021trivialaugment} and YOCO \cite{han2022you}. We refer to them as $T_{RA}$, $T_{TA}$ and $T_{YOCO}$. For $T_{YOCO}$, it may be based on $T_{RA}$ or $T_{TA}$. And we compare them with previous SOTA augmentations $T_{JO}$ and $T_{JC}$ for SSHPE, and our recommended $T_{JOCO}$ and $T_{JCCM}$. We did not compare to Mixup families \cite{zhang2018mixup, hendrycks2020augmix, pinto2022using, liu2022automix} or AutoAug families \cite{cubuk2018autoaugment, lim2019fast, hataya2020faster, zheng2022deep}. Because Mixup is verified not to work with SSHPE, and AutoAug needs to search optimal parameters. Finally, as shown in Tab.~\ref{tabOtherAugs} and Fig.~\ref{otherAugs}, our optimal combinations are always better than $T_{RA}$-based or $T_{TA}$-based $T_{YOCO}$, which are meticulously designed and also composed of existing basic augmentations. These further prove the superiority and conciseness of our synergistic combinations.

\begin{figure}[!]
        \centering
        \includegraphics[width=\columnwidth]{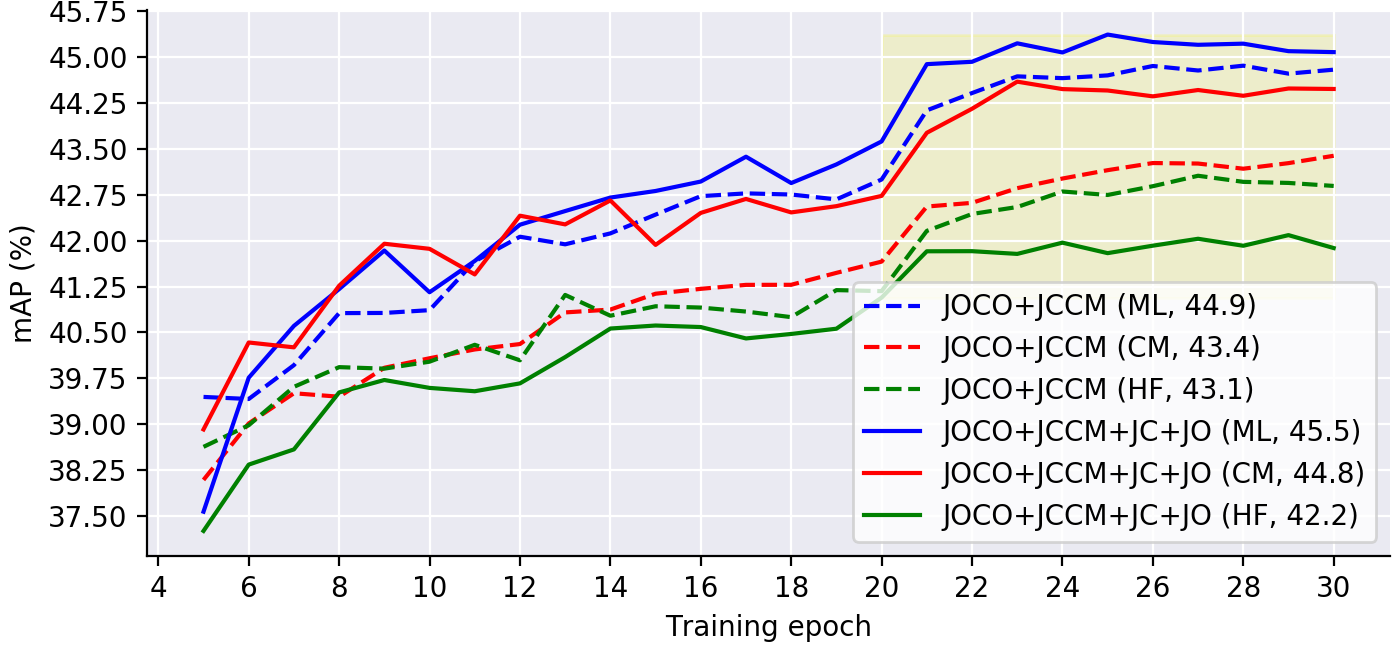}
        \vspace{-20pt}
        \caption{The convergence curves and best mAP results of two MultiAugs schemes when using different training techniques.}
        \label{MultiHeatsAS}
        \vspace{-10pt}
\end{figure}

\begin{figure}[!]
        \centering
        \includegraphics[width=\columnwidth]{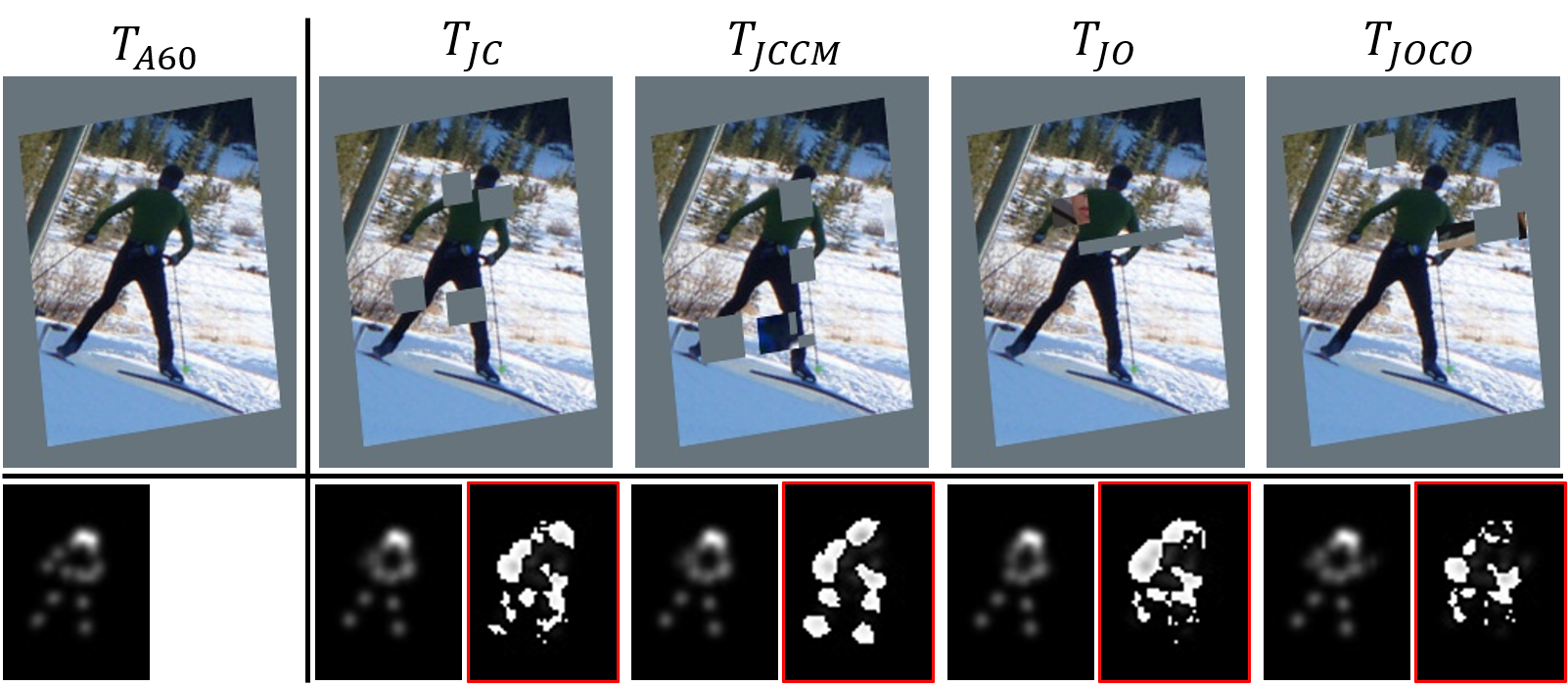}
        \vspace{-15pt}
        \caption{The predicted heatmaps of one easy augmentation ($T_{A60}$) and four hard augmentations in $m_{10}$. We also report the pixel-wise heatmap difference (with red borders) of each easy-hard pair to highlight subtle dissimilarities.}
        \label{MultiHeatsCase}
        \vspace{-10pt}
\end{figure}

\subsubsection{Training Techniques of Multiple Heatmaps}
In this part, we present additional techniques commonly used for unsupervised consistency training. Especially, for predicted multi-heatmaps, we propose to optimize them by applying the multi-loss learning (ML) as in Eq.~\ref{lossesunsup}. Other two alternative techniques are {\it confidence masking} (CM) and {\it heatmaps fusion} (HF). For CM, the consistency loss term in each mini-batch is computed only on keypoint channels whose maximum activation value is greater than a threshold $\tau$, which is set as 0.5. For HF, also termed as heatmaps ensemble, we sum and average multi-path heatmaps to obtain a fused heatmap for loss computing. Then, we compare MultiAugs (e.g., $m_{11}$ and $m_{12}$) using either of these three techniques ML, CM and HF. As shown in Fig.~\ref{MultiHeatsAS}, our ML is strictly superior than the other two techniques under either $m_{11}$ or $m_{12}$. For CM, we assume it may filter out some keypoint heatmaps with low confidence but high quality. This surely has a negative impact. For HF, although it is widely used in other SSL tasks for model ensemble, it may not necessarily be applicable to our intermediate keypoint heatmaps. We deem this is because each predicted heatmap is distinctive and meaningful (please refer Fig.~\ref{MultiHeatsCase} for details). It is tricky to replace them equivalently with a fused heatmap. In comparison, our multi-loss learning is a simple yet effective choice.

\begin{table}[t] 
	\centering
	\caption{AP results of baseline methods after increasing the batch size. The used backbone is ResNet18.}
	\vspace{-5pt}
	\label{tab10}
	\setlength{\tabcolsep}{4pt}
	\begin{tabular}{l|c|c|c|ccc}
	\Xhline{1.2pt}
	\multirow{2}{*}{Method} & \multirow{2}{*}{Nets} & \multirow{2}{*}{Loss Num.} & \multirow{2}{*}{Batch Size} & \multicolumn{3}{c}{Labeled Samples} \\
	\cline{5-7}
	~ & ~ & ~ & ~ & 1K & 5K & 10K \\
	\hline
	PoseCons & 1 & 1 & 32 & 42.1 & 52.3 & 57.3  \\
	PoseCons & 1 & 1 & 128 & 42.3 & 52.6 & 57.5  \\
	PoseDual & 2 & 1 & 32 & 44.6 & 55.6 & 59.6  \\
	PoseDual & 2 & 1 & 128 & 44.9 & 58.7 & 59.6  \\
	\Xhline{1.2pt}
	\end{tabular}
	\vspace{-5pt}
\end{table}

\begin{table}[t] 
	\centering
	\caption{AP results of our methods after adjusting the augmentation way of inputs. The used backbone is ResNet18.}
	\vspace{-5pt}
	\label{tab11}
	\setlength{\tabcolsep}{6pt}
	\begin{tabular}{l|c|c|ccc}
	\Xhline{1.2pt}
	\multirow{2}{*}{Method} & \multirow{2}{*}{Nets} & \multirow{2}{*}{Input Style} & \multicolumn{3}{c}{Labeled Samples} \\
	\cline{4-6}
	~ & ~ & ~ & 1K & 5K & 10K \\
	\hline
	Ours (Single) & 1 & 1-vs-n & 45.5 & 56.2 & 59.9  \\
	Ours (Single) & 1 & n-vs-n & 45.6 & 56.4 & 59.8  \\
	Ours (Dual) & 2 & 1-vs-n & 49.7 & 58.8 & 61.8  \\
	Ours (Dual) & 2 & n-vs-n & 49.7 & 58.9 & 61.9  \\
	\Xhline{1.2pt}
	\end{tabular}
	\vspace{-5pt}
\end{table}

\begin{table}[!]  
	\centering
	\caption{Quantitative comparison of training time. The integer with the marker \# means multi-path losses number. }
	\vspace{-5pt}
	\label{tab12}
	\setlength{\tabcolsep}{6pt}
	\begin{tabular}{l||c|c|c|c}
	\Xhline{1.2pt}
	Method & PoseCons & \makecell{Ours\\(Single,2\#)} & \makecell{Ours\\(Single,3\#)} & \makecell{Ours\\(Single,4\#)} \\
	\hline
	Time & $T_0$ & 1.36*$T_0$  & 1.50*$T_0$ & 1.83*$T_0$ \\
	\hline
	\hline
	Method & PoseDual &\makecell{Ours\\(Dual,2\#)} & \makecell{Ours\\(Dual,3\#)} & \makecell{Ours\\(Dual,4\#)} \\
	\hline
	Time & 2.49*$T_0$ & 2.62*$T_0$ & 2.88*$T_0$ & 3.14*$T_0$ \\
	\Xhline{1.2pt}
	\end{tabular}
	\vspace{-10pt}
\end{table}

\subsubsection{Influence of Batch Size}
We here investigate whether the multi-path consistency loss is sensitive to the training batch size. In fact, when performing ablation studies in Tab.~\ref{tabAugsPlusAS} using the single-path loss and Tab.~\ref{tabMultiAugsAS} using multi-path losses, we always set a fixed batch size 32. Moreover, for comparative experiments in Tab.~\ref{tab1}, we still keep the batch size as 32 and use the optimal four-path losses. Now, to investigate the possible impact of different batch sizes, we report the effects of PoseCons and PoseDual when the batch size is 128. As can be seen in Tab.~\ref{tab10}, after increasing the batch size of PoseCons or PoseDual accordingly, the final mAP results under different labeling rates (e.g., 1K, 5K and 10K) did not get significantly better. This indicates that batch size does not have a large impact on the performance of existing methods.

\subsubsection{Influence of Easy Augmentations}
We also need to probe whether to use a single constant easy augmentation as input for multi-path losses (the pair $\{I_e\}$ + $\{I_{h_1},...,I_{h_n}\}$, termed as 1-vs-n) or to use different easy augmentations multiple times as input (the pair $\{I_{e_1}$,...,$I_{e_n}\}$  + \{$I_{h_1}$,...,$I_{h_n}\}$, termed as n-vs-n). As shown in Tab.~\ref{tab11}, whether using 1-v-n augmented input or n-vs-n augmented input, the final mAP results obtained under various labeling rates are not significantly different. This is mainly because the used easy augmentation is always fixed (e.g., $T_{A30}$), so the input does not change in essence when applying 1-vs-n input or n-vs-n. These also reveal the necessity and rationality for us to focus on discussing how to find more advanced hard augmentations.

\subsubsection{Training Efficiency of Our Method}
Finally, in order to fairly and reasonably reflect the efficiency of our method in the training phase, we compare the training time of our method with that of PoseCons and PoseDual. The strong augmentation used by them is $T_{JC}$. Considering that our method often uses different strong augmentations, their computation is not the main bottleneck. Therefore, all strong augmentations in our method are also replaced into $T_{JC}$. Assuming that the total training time of PoseCons is one unit time $T_0$, which is actually about 7 hours. Then the total training time of running other methods is summarized in Tab.~\ref{tab12}. We can see that when using four-path losses, although the training time increases, it is still faster than PoseDual (1.83*$T_0$ vs. 2.49*$T_0$). Referring to the quantitative results in Tab.~\ref{tab1}, our method using a single-network and four-path losses achieves higher mAP than PoseDual. In addition, when using a dual-network with four-path losses, the total training time does not increase significantly (2.49*$T_0$  vs. 3.14*$T_0$). These indicate that our method is both efficient and effective.

\section{Conclusions}
In this paper, we aim to boost semi-supervised human pose estimation (SSHPE) from two perspectives: data augmentation and consistency training. Instead of inventing advanced augmentations in isolation, we attempt to synergistically utilize existing augmentations, and handily generate superior ones by novel combination paradigms. The discovered collaborative combinations have intuitive interpretability. We verified their advantages in solving the SSHPE problem. For consistency training, we abandon the convention of stacking networks to increase unsupervised losses, and train a single network by optimizing multi-path consistency losses for the same batch of unlabeled images. Combined with the optimal hard augmentations set, this plain and compact strategy is proven to be effective, and leads to better performance on various public benchmarks across different human keypoint domains. Last but not least, we declare that the synergy effects of augmentations and multiple consistency losses are generic and generalizable for other SSL vision tasks such as image classification, object detection, semantic segmentation and 3D regression. We will explore them in the future.




\bibliographystyle{IEEEtran}
\bibliography{refs_short}

\begin{thebibliography}{10}
\providecommand{\url}[1]{#1}
\csname url@samestyle\endcsname
\providecommand{\newblock}{\relax}
\providecommand{\bibinfo}[2]{#2}
\providecommand{\BIBentrySTDinterwordspacing}{\spaceskip=0pt\relax}
\providecommand{\BIBentryALTinterwordstretchfactor}{4}
\providecommand{\BIBentryALTinterwordspacing}{\spaceskip=\fontdimen2\font plus
\BIBentryALTinterwordstretchfactor\fontdimen3\font minus
  \fontdimen4\font\relax}
\providecommand{\BIBforeignlanguage}[2]{{%
\expandafter\ifx\csname l@#1\endcsname\relax
\typeout{** WARNING: IEEEtran.bst: No hyphenation pattern has been}%
\typeout{** loaded for the language `#1'. Using the pattern for}%
\typeout{** the default language instead.}%
\else
\language=\csname l@#1\endcsname
\fi
#2}}
\providecommand{\BIBdecl}{\relax}
\BIBdecl

\bibitem{yan2018spatial}
S.~Yan, Y.~Xiong, and D.~Lin, ``Spatial temporal graph convolutional networks
  for skeleton-based action recognition,'' in \emph{AAAI}, vol.~32, no.~1,
  2018.

\bibitem{duan2022revisiting}
H.~Duan, Y.~Zhao, K.~Chen, D.~Lin, and B.~Dai, ``Revisiting skeleton-based
  action recognition,'' in \emph{CVPR}, 2022, pp. 2969--2978.

\bibitem{zhao2017spindle}
H.~Zhao, M.~Tian, S.~Sun, J.~Shao, J.~Yan, S.~Yi, X.~Wang, and X.~Tang,
  ``Spindle net: Person re-identification with human body region guided feature
  decomposition and fusion,'' in \emph{CVPR}, 2017, pp. 1077--1085.

\bibitem{sarfraz2018pose}
M.~S. Sarfraz, A.~Schumann, A.~Eberle, and R.~Stiefelhagen, ``A pose-sensitive
  embedding for person re-identification with expanded cross neighborhood
  re-ranking,'' in \emph{CVPR}, 2018, pp. 420--429.

\bibitem{nie2023lifting}
Q.~Nie, Z.~Liu, and Y.~Liu, ``Lifting 2d human pose to 3d with domain adapted
  3d body concept,'' \emph{IJCV}, vol. 131, no.~5, pp. 1250--1268, 2023.

\bibitem{dabhi20243d}
M.~Dabhi, L.~A. Jeni, and S.~Lucey, ``3d-lfm: Lifting foundation model,'' in
  \emph{CVPR}, 2024, pp. 10\,466--10\,475.

\bibitem{pavlakos2018learning}
G.~Pavlakos, L.~Zhu, X.~Zhou, and K.~Daniilidis, ``Learning to estimate 3d
  human pose and shape from a single color image,'' in \emph{CVPR}, 2018, pp.
  459--468.

\bibitem{pavlakos2019expressive}
G.~Pavlakos, V.~Choutas, N.~Ghorbani, T.~Bolkart, A.~A. Osman, D.~Tzionas, and
  M.~J. Black, ``Expressive body capture: 3d hands, face, and body from a
  single image,'' in \emph{CVPR}, 2019, pp. 10\,975--10\,985.

\bibitem{cao2017realtime}
Z.~Cao, T.~Simon, S.-E. Wei, and Y.~Sheikh, ``Realtime multi-person 2d pose
  estimation using part affinity fields,'' in \emph{CVPR}, 2017, pp.
  7291--7299.

\bibitem{cheng2020higherhrnet}
B.~Cheng, B.~Xiao, J.~Wang, H.~Shi, T.~S. Huang, and L.~Zhang, ``Higherhrnet:
  Scale-aware representation learning for bottom-up human pose estimation,'' in
  \emph{CVPR}, 2020, pp. 5386--5395.

\bibitem{xu2022vitpose}
Y.~Xu, J.~Zhang, Q.~Zhang, and D.~Tao, ``Vitpose: Simple vision transformer
  baselines for human pose estimation,'' \emph{NeurIPS}, vol.~35, pp.
  38\,571--38\,584, 2022.

\bibitem{yang2023explicit}
J.~Yang, A.~Zeng, S.~Liu, F.~Li, R.~Zhang, and L.~Zhang, ``Explicit box
  detection unifies end-to-end multi-person pose estimation,'' \emph{ICLR},
  2023.

\bibitem{tan2024diffusionregpose}
D.~Tan, H.~Chen, W.~Tian, and L.~Xiong, ``Diffusionregpose: Enhancing
  multi-person pose estimation using a diffusion-based end-to-end regression
  approach,'' in \emph{CVPR}, 2024, pp. 2230--2239.

\bibitem{andriluka20142d}
M.~Andriluka, L.~Pishchulin, P.~Gehler, and B.~Schiele, ``2d human pose
  estimation: New benchmark and state of the art analysis,'' in \emph{CVPR},
  2014, pp. 3686--3693.

\bibitem{lin2014microsoft}
T.-Y. Lin, M.~Maire, S.~Belongie, J.~Hays, P.~Perona, D.~Ramanan,
  P.~Doll{\'a}r, and C.~L. Zitnick, ``Microsoft coco: Common objects in
  context,'' in \emph{ECCV}.\hskip 1em plus 0.5em minus 0.4em\relax Springer,
  2014, pp. 740--755.

\bibitem{wu2019large}
J.~Wu, H.~Zheng, B.~Zhao, Y.~Li, B.~Yan, R.~Liang, W.~Wang, S.~Zhou, G.~Lin,
  Y.~Fu \emph{et~al.}, ``Large-scale datasets for going deeper in image
  understanding,'' in \emph{ICME}.\hskip 1em plus 0.5em minus 0.4em\relax IEEE,
  2019, pp. 1480--1485.

\bibitem{xie2021empirical}
R.~Xie, C.~Wang, W.~Zeng, and Y.~Wang, ``An empirical study of the collapsing
  problem in semi-supervised 2d human pose estimation,'' in \emph{ICCV}.

\bibitem{moskvyak2021semi}
O.~Moskvyak, F.~Maire, F.~Dayoub, and M.~Baktashmotlagh, ``Semi-supervised
  keypoint localization,'' in \emph{ICLR}, 2021.

\bibitem{wang2022pseudo}
C.~Wang, S.~Jin, Y.~Guan, W.~Liu, C.~Qian, P.~Luo, and W.~Ouyang,
  ``Pseudo-labeled auto-curriculum learning for semi-supervised keypoint
  localization,'' in \emph{ICLR}, 2022.

\bibitem{huang2023semi}
L.~Huang, Y.~Li, H.~Tian, Y.~Yang, X.~Li, W.~Deng, and J.~Ye, ``Semi-supervised
  2d human pose estimation driven by position inconsistency pseudo label
  correction module,'' in \emph{CVPR}, 2023, pp. 693--703.

\bibitem{yu2024denoising}
Z.~Yu, M.~Wang, Y.~Chen, P.~Favaro, and D.~Modolo, ``Denoising and selecting
  pseudo-heatmaps for semi-supervised human pose estimation,'' in \emph{WACV},
  2024, pp. 6280--6289.

\bibitem{devries2017improved}
T.~DeVries and G.~W. Taylor, ``Improved regularization of convolutional neural
  networks with cutout,'' \emph{arXiv preprint arXiv:1708.04552}, 2017.

\bibitem{zhang2018mixup}
H.~Zhang, M.~Cisse, Y.~N. Dauphin, and D.~Lopez-Paz, ``mixup: Beyond empirical
  risk minimization,'' in \emph{ICLR}, 2018.

\bibitem{yun2019cutmix}
S.~Yun, D.~Han, S.~J. Oh, S.~Chun, J.~Choe, and Y.~Yoo, ``Cutmix:
  Regularization strategy to train strong classifiers with localizable
  features,'' in \emph{ICCV}, 2019, pp. 6023--6032.

\bibitem{cubuk2018autoaugment}
E.~D. Cubuk, B.~Zoph, D.~Mane, V.~Vasudevan, and Q.~V. Le, ``Autoaugment:
  Learning augmentation policies from data,'' \emph{arXiv preprint
  arXiv:1805.09501}, 2018.

\bibitem{cubuk2020randaugment}
E.~D. Cubuk, B.~Zoph, J.~Shlens, and Q.~V. Le, ``Randaugment: Practical
  automated data augmentation with a reduced search space,'' in \emph{CVPRW},
  2020, pp. 702--703.

\bibitem{lim2019fast}
S.~Lim, I.~Kim, T.~Kim, C.~Kim, and S.~Kim, ``Fast autoaugment,''
  \emph{NeurIPS}, vol.~32, 2019.

\bibitem{hataya2020faster}
R.~Hataya, J.~Zdenek, K.~Yoshizoe, and H.~Nakayama, ``Faster autoaugment:
  Learning augmentation strategies using backpropagation,'' in
  \emph{ECCV}.\hskip 1em plus 0.5em minus 0.4em\relax Springer, 2020, pp.
  1--16.

\bibitem{zheng2022deep}
Y.~Zheng, Z.~Zhang, S.~Yan, and M.~Zhang, ``Deep autoaugment,'' in \emph{ICLR},
  2022.

\bibitem{hendrycks2020augmix}
D.~Hendrycks, N.~Mu, E.~D. Cubuk, B.~Zoph, J.~Gilmer, and B.~Lakshminarayanan,
  ``Augmix: A simple data processing method to improve robustness and
  uncertainty,'' in \emph{ICLR}, 2020.

\bibitem{muller2021trivialaugment}
S.~G. M{\"u}ller and F.~Hutter, ``Trivialaugment: Tuning-free yet
  state-of-the-art data augmentation,'' in \emph{ICCV}, 2021, pp. 774--782.

\bibitem{pinto2022using}
F.~Pinto, H.~Yang, S.~N. Lim, P.~Torr, and P.~Dokania, ``Using mixup as a
  regularizer can surprisingly improve accuracy \& out-of-distribution
  robustness,'' \emph{NeurIPS}, vol.~35, pp. 14\,608--14\,622, 2022.

\bibitem{liu2022automix}
Z.~Liu, S.~Li, D.~Wu, Z.~Liu, Z.~Chen, L.~Wu, and S.~Z. Li, ``Automix:
  Unveiling the power of mixup for stronger classifiers,'' in
  \emph{ECCV}.\hskip 1em plus 0.5em minus 0.4em\relax Springer, 2022, pp.
  441--458.

\bibitem{han2022you}
J.~Han, P.~Fang, W.~Li, J.~Hong, M.~A. Armin, I.~Reid, L.~Petersson, and H.~Li,
  ``You only cut once: Boosting data augmentation with a single cut,'' in
  \emph{ICML}.\hskip 1em plus 0.5em minus 0.4em\relax PMLR, 2022, pp.
  8196--8212.

\bibitem{radosavovic2018data}
I.~Radosavovic, P.~Doll{\'a}r, R.~Girshick, G.~Gkioxari, and K.~He, ``Data
  distillation: Towards omni-supervised learning,'' in \emph{CVPR}, 2018, pp.
  4119--4128.

\bibitem{xie2020unsupervised}
Q.~Xie, Z.~Dai, E.~Hovy, T.~Luong, and Q.~Le, ``Unsupervised data augmentation
  for consistency training,'' \emph{NeurIPS}, vol.~33, pp. 6256--6268, 2020.

\bibitem{oliver2018realistic}
A.~Oliver, A.~Odena, C.~A. Raffel, E.~D. Cubuk, and I.~Goodfellow, ``Realistic
  evaluation of deep semi-supervised learning algorithms,'' \emph{NeurIPS},
  vol.~31, 2018.

\bibitem{xie2020self}
Q.~Xie, M.-T. Luong, E.~Hovy, and Q.~V. Le, ``Self-training with noisy student
  improves imagenet classification,'' in \emph{CVPR}, 2020, pp.
  10\,687--10\,698.

\bibitem{sohn2020fixmatch}
K.~Sohn, D.~Berthelot, N.~Carlini, Z.~Zhang, H.~Zhang, C.~A. Raffel, E.~D.
  Cubuk, A.~Kurakin, and C.-L. Li, ``Fixmatch: Simplifying semi-supervised
  learning with consistency and confidence,'' \emph{NeurIPS}, vol.~33, pp.
  596--608, 2020.

\bibitem{guo2022class}
L.-Z. Guo and Y.-F. Li, ``Class-imbalanced semi-supervised learning with
  adaptive thresholding,'' in \emph{ICML}.\hskip 1em plus 0.5em minus
  0.4em\relax PMLR, 2022, pp. 8082--8094.

\bibitem{wang2023freematch}
Y.~Wang, H.~Chen, Q.~Heng, W.~Hou, Y.~Fan, Z.~Wu, J.~Wang, M.~Savvides,
  T.~Shinozaki, B.~Raj \emph{et~al.}, ``Freematch: Self-adaptive thresholding
  for semi-supervised learning,'' in \emph{ICLR}, 2023.

\bibitem{laine2016temporal}
S.~Laine and T.~Aila, ``Temporal ensembling for semi-supervised learning,'' in
  \emph{ICLR}, 2016.

\bibitem{tarvainen2017mean}
A.~Tarvainen and H.~Valpola, ``Mean teachers are better role models:
  Weight-averaged consistency targets improve semi-supervised deep learning
  results,'' \emph{NeurIPS}, vol.~30, 2017.

\bibitem{berthelot2019mixmatch}
D.~Berthelot, N.~Carlini, I.~Goodfellow, N.~Papernot, A.~Oliver, and C.~A.
  Raffel, ``Mixmatch: A holistic approach to semi-supervised learning,''
  \emph{NeurIPS}, vol.~32, 2019.

\bibitem{zhang2021flexmatch}
B.~Zhang, Y.~Wang, W.~Hou, H.~Wu, J.~Wang, M.~Okumura, and T.~Shinozaki,
  ``Flexmatch: Boosting semi-supervised learning with curriculum pseudo
  labeling,'' \emph{NeurIPS}, vol.~34, pp. 18\,408--18\,419, 2021.

\bibitem{gui2023enhancing}
G.~Gui, Z.~Zhao, L.~Qi, L.~Zhou, L.~Wang, and Y.~Shi, ``Enhancing sample
  utilization through sample adaptive augmentation in semi-supervised
  learning,'' in \emph{ICCV}, 2023, pp. 15\,880--15\,889.

\bibitem{huang2024interlude}
Z.~Huang, X.~Yu, D.~Zhu, and M.~C. Hughes, ``Interlude: Interactions between
  labeled and unlabeled data to enhance semi-supervised learning,'' in
  \emph{ICML}.\hskip 1em plus 0.5em minus 0.4em\relax PMLR, 2024.

\bibitem{fan2023revisiting}
Y.~Fan, A.~Kukleva, D.~Dai, and B.~Schiele, ``Revisiting consistency
  regularization for semi-supervised learning,'' \emph{IJCV}, vol. 131, no.~3,
  pp. 626--643, 2023.

\bibitem{springstein2022semi}
M.~Springstein, S.~Schneider, C.~Althaus, and R.~Ewerth, ``Semi-supervised
  human pose estimation in art-historical images,'' in \emph{ACMMM}, 2022, pp.
  1107--1116.

\bibitem{li2023scarcenet}
C.~Li and G.~H. Lee, ``Scarcenet: Animal pose estimation with scarce
  annotations,'' in \emph{CVPR}, 2023, pp. 17\,174--17\,183.

\bibitem{han2022learning}
Z.~Han, H.~Sun, and Y.~Yin, ``Learning transferable parameters for unsupervised
  domain adaptation,'' \emph{TIP}, vol.~31, pp. 6424--6439, 2022.

\bibitem{johnson2011learning}
S.~Johnson and M.~Everingham, ``Learning effective human pose estimation from
  inaccurate annotation,'' in \emph{CVPR}, 2011, pp. 1465--1472.

\bibitem{qiao2018deep}
S.~Qiao, W.~Shen, Z.~Zhang, B.~Wang, and A.~Yuille, ``Deep co-training for
  semi-supervised image recognition,'' in \emph{ECCV}, 2018, pp. 135--152.

\bibitem{ke2019dual}
Z.~Ke, D.~Wang, Q.~Yan, J.~Ren, and R.~W. Lau, ``Dual student: Breaking the
  limits of the teacher in semi-supervised learning,'' in \emph{ICCV}, 2019,
  pp. 6728--6736.

\bibitem{xiao2018simple}
B.~Xiao, H.~Wu, and Y.~Wei, ``Simple baselines for human pose estimation and
  tracking,'' in \emph{ECCV}, 2018, pp. 466--481.

\bibitem{sun2019deep}
K.~Sun, B.~Xiao, D.~Liu, and J.~Wang, ``Deep high-resolution representation
  learning for human pose estimation,'' in \emph{CVPR}, 2019, pp. 5693--5703.

\bibitem{caron2020unsupervised}
M.~Caron, I.~Misra, J.~Mairal, P.~Goyal, P.~Bojanowski, and A.~Joulin,
  ``Unsupervised learning of visual features by contrasting cluster
  assignments,'' \emph{NeurIPS}, vol.~33, pp. 9912--9924, 2020.

\bibitem{berthelot2020remixmatch}
D.~Berthelot, N.~Carlini, E.~D. Cubuk, A.~Kurakin, K.~Sohn, H.~Zhang, and
  C.~Raffel, ``Remixmatch: Semi-supervised learning with distribution matching
  and augmentation anchoring,'' in \emph{ICLR}, 2020.

\bibitem{chen2020simple}
T.~Chen, S.~Kornblith, M.~Norouzi, and G.~Hinton, ``A simple framework for
  contrastive learning of visual representations,'' in \emph{ICML}.\hskip 1em
  plus 0.5em minus 0.4em\relax PMLR, 2020, pp. 1597--1607.

\bibitem{he2020momentum}
K.~He, H.~Fan, Y.~Wu, S.~Xie, and R.~Girshick, ``Momentum contrast for
  unsupervised visual representation learning,'' in \emph{CVPR}, 2020, pp.
  9729--9738.

\bibitem{chen2021exploring}
X.~Chen and K.~He, ``Exploring simple siamese representation learning,'' in
  \emph{CVPR}, 2021, pp. 15\,750--15\,758.

\bibitem{li2021comatch}
J.~Li, C.~Xiong, and S.~C. Hoi, ``Comatch: Semi-supervised learning with
  contrastive graph regularization,'' in \emph{ICCV}, 2021, pp. 9475--9484.

\bibitem{yang2022class}
F.~Yang, K.~Wu, S.~Zhang, G.~Jiang, Y.~Liu, F.~Zheng, W.~Zhang, C.~Wang, and
  L.~Zeng, ``Class-aware contrastive semi-supervised learning,'' in
  \emph{CVPR}, 2022, pp. 14\,421--14\,430.

\bibitem{wu2023chmatch}
J.~Wu, H.~Yang, T.~Gan, N.~Ding, F.~Jiang, and L.~Nie, ``Chmatch: Contrastive
  hierarchical matching and robust adaptive threshold boosted semi-supervised
  learning,'' in \emph{CVPR}, 2023, pp. 15\,762--15\,772.

\bibitem{chen2019transferability}
X.~Chen, S.~Wang, M.~Long, and J.~Wang, ``Transferability vs. discriminability:
  Batch spectral penalization for adversarial domain adaptation,'' in
  \emph{ICML}.\hskip 1em plus 0.5em minus 0.4em\relax PMLR, 2019, pp.
  1081--1090.

\bibitem{xue2022investigating}
Y.~Xue, K.~Whitecross, and B.~Mirzasoleiman, ``Investigating why contrastive
  learning benefits robustness against label noise,'' in \emph{ICML}.\hskip 1em
  plus 0.5em minus 0.4em\relax PMLR, 2022, pp. 24\,851--24\,871.

\bibitem{duan2020rapid}
Z.~Duan, O.~Tezcan, H.~Nakamura, P.~Ishwar, and J.~Konrad, ``Rapid:
  rotation-aware people detection in overhead fisheye images,'' in
  \emph{CVPRW}, 2020, pp. 636--637.

\bibitem{tezcan2022wepdtof}
O.~Tezcan, Z.~Duan, M.~Cokbas, P.~Ishwar, and J.~Konrad, ``Wepdtof: A dataset
  and benchmark algorithms for in-the-wild people detection and tracking from
  overhead fisheye cameras,'' in \emph{WACV}, 2022, pp. 503--512.

\bibitem{jin2020whole}
S.~Jin, L.~Xu, J.~Xu, C.~Wang, W.~Liu, C.~Qian, W.~Ouyang, and P.~Luo,
  ``Whole-body human pose estimation in the wild,'' in \emph{ECCV}.\hskip 1em
  plus 0.5em minus 0.4em\relax Springer, 2020, pp. 196--214.

\bibitem{zhou2024bpjdet}
H.~Zhou, F.~Jiang, J.~Si, Y.~Ding, and H.~Lu, ``Bpjdet: Extended object
  representation for generic body-part joint detection,'' \emph{TPAMI}, 2024.

\bibitem{he2016deep}
K.~He, X.~Zhang, S.~Ren, and J.~Sun, ``Deep residual learning for image
  recognition,'' in \emph{CVPR}, 2016, pp. 770--778.

\bibitem{yang2021transpose}
S.~Yang, Z.~Quan, M.~Nie, and W.~Yang, ``Transpose: Keypoint localization via
  transformer,'' in \emph{ICCV}, 2021, pp. 11\,802--11\,812.

\bibitem{li2021tokenpose}
Y.~Li, S.~Zhang, Z.~Wang, S.~Yang, W.~Yang, S.-T. Xia, and E.~Zhou,
  ``Tokenpose: Learning keypoint tokens for human pose estimation,'' in
  \emph{ICCV}, 2021, pp. 11\,313--11\,322.

\bibitem{yuan2021hrformer}
Y.~Yuan, R.~Fu, L.~Huang, W.~Lin, C.~Zhang, X.~Chen, and J.~Wang, ``Hrformer:
  High-resolution transformer for dense prediction,'' \emph{NeurIPS}, vol.~34,
  pp. 7281--7293, 2021.

\bibitem{li2019rethinking}
W.~Li, Z.~Wang, B.~Yin, Q.~Peng, Y.~Du, T.~Xiao, G.~Yu, H.~Lu, Y.~Wei, and
  J.~Sun, ``Rethinking on multi-stage networks for human pose estimation,''
  \emph{arXiv preprint arXiv:1901.00148}, 2019.

\bibitem{zhang2020distribution}
F.~Zhang, X.~Zhu, H.~Dai, M.~Ye, and C.~Zhu, ``Distribution-aware coordinate
  representation for human pose estimation,'' in \emph{CVPR}, 2020, pp.
  7093--7102.

\bibitem{huang2020devil}
J.~Huang, Z.~Zhu, F.~Guo, and G.~Huang, ``The devil is in the details: Delving
  into unbiased data processing for human pose estimation,'' in \emph{CVPR},
  2020, pp. 5700--5709.

\bibitem{newell2016stacked}
A.~Newell, K.~Yang, and J.~Deng, ``Stacked hourglass networks for human pose
  estimation,'' in \emph{ECCV}.\hskip 1em plus 0.5em minus 0.4em\relax
  Springer, 2016, pp. 483--499.

\bibitem{ke2018multi}
L.~Ke, M.-C. Chang, H.~Qi, and S.~Lyu, ``Multi-scale structure-aware network
  for human pose estimation,'' in \emph{ECCV}, 2018, pp. 713--728.

\bibitem{zhang2019human}
H.~Zhang, H.~Ouyang, S.~Liu, X.~Qi, X.~Shen, R.~Yang, and J.~Jia, ``Human pose
  estimation with spatial contextual information,'' \emph{arXiv preprint
  arXiv:1901.01760}, 2019.

\end{thebibliography}

\end{document}